\documentclass[twoside,11pt]{article}
\usepackage{blindtext}

\usepackage[preprint]{jmlr2e}

\usepackage{lastpage}

\firstpageno{1}

\usepackage{amsmath}
\usepackage{amssymb}
\usepackage{amsthm} 
\usepackage{abbreviations}
\usepackage{natbib}
    \setcitestyle{authoryear,round,citesep={;},aysep={,},yysep={;}}
    \renewcommand{\cite}[1]{\citep{#1}}
    
\usepackage{hyperref}
    \hypersetup{ %
        pdftitle={},
        pdfsubject={},
        pdfkeywords={},
        pdfborder=0 0 0,
        pdfpagemode=UseNone,
        colorlinks=true,
        linkcolor=black,
        citecolor=mydarkblue, 
        filecolor=mydarkblue,
        urlcolor=mydarkblue,
    }

\usepackage{algorithm}
\usepackage{algorithmic}
\usepackage{subcaption}
\usepackage{booktabs}       

\usepackage{tikz} 
    \usetikzlibrary{shapes,arrows,positioning}

\addtocontents{toc}{\protect\setcounter{tocdepth}{0}} 

\begin{document}

\title{Model-Based Reinforcement Learning with Multinomial Logistic Function Approximation}

\author{\name Taehyun Hwang \email th.hwang@snu.ac.kr \\
       \addr Graduate School of Data Science\\
       Seoul National University\\
       Seoul, Republic of Korea
       \AND
       \name Min-hwan Oh\thanks{The corresponding author} \email minoh@snu.ac.kr \\ 
       \addr Graduate School of Data Science\\
       Seoul National University\\
       Seoul, Republic of Korea}

\editor{My editor}

\maketitle

\begin{abstract}%
We study model-based reinforcement learning (RL) for episodic Markov decision processes (MDP) whose transition probability is parametrized by an unknown transition core with features of state and action. Despite much recent progress in analyzing algorithms in the linear MDP setting, the understanding of more general transition models is very restrictive. In this paper, we establish a provably efficient RL algorithm for the MDP whose state transition is given by a multinomial logistic model. To balance the exploration-exploitation trade-off, we propose an upper confidence bound-based algorithm.
We show that our proposed algorithm achieves $\tilde{\mathcal{O}}(d \sqrt{H^3 T})$ regret bound where $d$ is the dimension of the transition core, $H$ is the horizon, and $T$ is the total number of steps. To the best of our knowledge, this is the first model-based RL algorithm with multinomial logistic function approximation with provable guarantees. We also comprehensively evaluate our proposed algorithm numerically and show that it consistently outperforms the existing methods, hence achieving both provable efficiency and practical superior performance.
\end{abstract}


\section{Introduction}
Reinforcement learning (RL) with function approximation has made significant advances in empirical studies~\citet{mnih2015human, silver2017mastering, silver2018general}. However, the theoretical understanding of these methods is still limited.
Recently, function approximation with provable efficiency has been gaining significant attention in the research community, trying to close the gap between theory and empirical findings.
Most of the existing theoretical works in RL with function approximation consider linear function approximation~\citet{jiang2017contextual, yang2019sample,  yang2020reinforcement, jin2020provably, zanette2020frequentist, modi2020sample, du2020is, cai2020provably, ayoub2020model, wang2020reinforcement_eluder, weisz2021exponential, he2021logarithmic, zhou2021nearly, zhou2021provably, ishfaq2021randomized}.
Many of these linear model-based methods and their analyses rely on the classical upper confidence bound (UCB) or randomized exploration methods such as Thompson sampling extending the analysis of linear contextual bandits \citet{chu2011contextual,abbasi2011improved, agrawal2013thompson, abeille2017Linear, kveton2020perturbed}. 

While new methods are still being proposed under the linearity assumption and performance guarantees have been improved, the linear model assumption on the transition model of Markov decision processes (MDPs) faces a simple yet fundamental challenge. A transition model in MDPs is a \textit{probability distribution} over states. A linear function approximating the transition model needs to satisfy that the function output is within $[0,1]$ and, furthermore, the probabilities over all possible next states sum to 1 exactly. Note that such requirements are not just imposed approximately, but rather exactly, since almost all existing works in linear function approximation assume realizability, i.e., the true transition model is assumed to be linear 
\citet{yang2020reinforcement, jin2020provably, zanette2020frequentist, zhou2021nearly, zhou2021provably,
ishfaq2021randomized}.

The linear model assumption also limits the set of feature representations of states or state-action pairs that are admissible for the transition model. In function approximation settings, the transition models are typically functions of feature representations. However, for a given linear transition model, an arbitrary feature may not induce a proper probability distribution. Put differently, nature can reveal a set of feature representations such that no linear model can properly construct a probability distribution over states.
Hence, the fundamental condition required for the true transition model can easily be violated for the linear model. This issue becomes even more challenging for a \textit{estimated} model.\footnote{That is, even if the true model is truly linear and satisfies the basic conditions for a probability distribution, the estimated model can still violate them.}
Furthermore, when there is model misspecification, sublinear guarantees on the regret performances that the existing methods enjoy become not valid, hence potentially leading to serious deterioration of the performances.

In supervised learning paradigm, a distribution over multiple possible outcomes is rarely learned using a separate linear model for each outcome.
For example, consider a learning problem with a binary outcome. One of the most obvious choices of a model to use in such a learning problem is a \textit{logistic} model.
Acknowledging that the state transition model of MDPs with a finite number of next states (the total number of states can be infinite) is essentially a categorical distribution, the \textit{multinomial logistic} (MNL) model is certainly one of the first choices to consider.
In statistics and machine learning research, the generalization of the linear model to a function class suitable for particular problem settings has been an important milestone both in terms of applicability and theoretical perspectives.
In parametric bandit research, a closely related field of RL, the extension of linear bandits to generalized linear bandits, including logistic bandits for binary feedback and multinomial logistic bandits for multiclass feedback, has been an active area of research~\citet{filippi2010parametric, li2017provably, jun2017scalable, kveton2020randomized, oh2019thompson, oh2021multinomial}. 
Surprisingly, there has been no prior work on RL with \textit{multinomial logistic} function approximation (or even logistic function approximation), in spite of a vast amount of literature on linear function approximation and despite the fact that the multinomial logistic model can naturally capture the state transition probabilities.

To the best of our knowledge, our work is the first to study a provably efficient RL under the multinomial logistic function approximation.
The generalization of the transition probability model beyond the simple linear model to the multinomial logistic model allows for broader applicability, overcoming the crucial limitations of the linear transition model (See Section~\ref{sec:linear_vs_MNL}).
On theoretical perspectives,
going beyond the linear model to the MNL model requires more involved analysis without closed-form solutions for the estimation 
and accounting for non-linearity.
Note that the linear model assumption for the transition model induces a linear value function which enables the utilization of the least-square estimation and the linear bandit techniques for regret analysis~\citet{jin2020provably,zanette2020frequentist,ishfaq2021randomized}.
However, in the MNL function approximation, we no longer have a linearly parametrized value function nor do we have any closed form expression for the value function.
It appears that these aspects pose greater technical challenges in RL with MNL function approximation.
Therefore, the following research question arises:

\textit{Can we design a provably efficient RL algorithm for the multinomial logistic transition model?}

In this paper, we address the above question affirmatively.
We study a finite-horizon RL problem where the transition probability is assumed to be a MNL model parametrized by a transition core.
We propose a provably efficient model-based RL algorithm that balances the exploration-exploitation trade-off, establishing the first results for the MNL transition model approximation of MDPs.
Our main contributions are summarized as follows:
\begin{itemize}

    \item 
    We formally discuss the shortcomings of the linear function approximation (e.g., Proposition~\ref{prop:arbitrary linear transition model} in Section 2.3). To the best of our knowledge, the rigorous discussion on the limitation of the linear transition model provides meaningful insights and may be of independent interest. The message here is that the linear transition model is restricted not just in the functional form, but also the set of features that satisfy the requirement imposed by the linear MDP is very limited. 

    \item 
    The MNL function approximation that we study in this paper is a much more flexible and practical function approximation than the linear function approximation which had been studied extensively in the recent literature. To our best knowledge, our paper is the first work to consider multinomial logistic function approximation (that provides provable guarantees) and hence, we believe, serves as an important milestone. We believe such a modeling assumption not only naturally captures the essence of the state transition probabilities, overcoming the drawbacks of the linear function approximation, but also induces an efficient algorithm that utilizes the structure. Moreover, we believe that the MNL function approximation will have a further impact on future research, paving the way for more results in this direction.

    \item We propose a provably efficient algorithm for model-based RL in feature space, Upper Confidence RL with MNL transition model ($\AlgName$).
    To the best of our knowledge, this is the first \textit{model-based} RL algorithm with
    multinomial logistic function approximation.
    \item We establish that $\AlgName$ is statistically efficient achieving $\tilde{\Ocal}(d \sqrt{H^3 T})$ regret, where $d$ is the dimension of the transition core, $H$ is the planning horizon, and $T$ is the total number of steps. 
    Noting that $d$ is the total dimension of the unknown parameter, 
    the dependence on dimensionality as well as dependence on total steps matches the corresponding dependence of the regret bound in linear MDPs~\citet{zhou2021nearly}.
    
    \item We evaluate our algorithm on numerical experiments and show that it consistently outperforms the existing provably efficient RL methods by significant margins. 
    We performed experiments on tabular MDPs. 
    Hence, no modeling assumption on the true functional form is imposed for the transition model, 
    which does not favor any particular model of approximation. The experiments provide the evidences that our proposed algorithm is both provably and practically efficient. 
\end{itemize}

The MNL function approximation that we study in this paper is a much more flexible and practical generalization of the tabular RL than linearly parametrized MDPs, which have been widely studied in the recent literature. As the first work to study RL with MNL transition model, we believe that both our proposed transition model and the proposed algorithm provide sound contributions in terms of theory and practicality.

\subsection{Related Work}\label{sec:related_work}

For tabular MDPs with a finite $H$-horizon, there are a large number of works both on model-based methods \citet{jaksch2010near, osband2014modelbased, azar2017minimax, dann2017unifying, agrawal2017posterior, ouyang2017learning} and on model-free methods \citet{jin2018qlearning, osband2019deep, russo2019worst, zhang2020almostoptimal, zhang2021modelfree}. Both model-based and model-free methods are known to achieve $\tilde{\Ocal}(H\sqrt{SAT})$ regret, where $S$ is the number of states, and $A$ is the number of actions. This bound is proven to be optimal up to logarithmic factors~\citet{jin2018qlearning, zhang2020almostoptimal}. 

Extending beyond tabular MDPs, there have been an increasing number of works on function approximation with provable guarantees \citet{jiang2017contextual, yang2019sample,  yang2020reinforcement, jin2020provably, zanette2020frequentist, modi2020sample, du2020is, cai2020provably, ayoub2020model, wang2020reinforcement_eluder, weisz2021exponential, he2021logarithmic, zhou2021nearly, zhou2021provably, ishfaq2021randomized}.
For regret minimization in RL with linear function approximation,
\citet{jin2020provably} assume that the transition model and the reward function of the MDPs are linear functions of a $d$-dimensional feature mapping and propose an optimistic
variant of the Least-squares Value Iteration (LSVI) algorithm 
\citet{bradtke1996linear, osband2016generalization}  with $\tilde{\Ocal}(d^{3/2}H^{3/2}\sqrt{T})$ regret. 
\citet{zanette2020frequentist} propose a randomized LSVI algorithm where exploration is induced by perturbing the least-squares approximation of the action-value function and provide $\tilde{\Ocal}(d^2H^2\sqrt{T})$ regret. 
For model-based methods with function approximation, \citet{yang2020reinforcement} assume the transition probability kernel to be a bilinear model  parametrized by a matrix and propose a model-based algorithm with $\tilde{\Ocal}(d^{3/2} H^2 \sqrt{T})$ regret. 
\citet{jia2020model} consider a special class of MDPs called linear mixture MDPs where the transition probability kernel is a linear mixture of a number of basis kernels, which covers various classes of MDPs studied in previous works \citet{modi2020sample, yang2020reinforcement}. For this model, \citet{jia2020model} propose a UCB-based RL algorithm with value-targeted model parameter estimation with $\tilde{\Ocal}(d H^{3/2} \sqrt{T})$ regret.
The same linear mixture MDPs has been also used by \citet{ayoub2020model,zhou2021nearly, zhou2021provably}. 
In particular, \citet{zhou2021nearly} propose a variant of the method proposed by \citet{jia2020model} and prove $\tilde{\Ocal}(d H \sqrt{T})$ regret with a matching lower bound $\Omega(d H \sqrt{T})$ for linear mixture MDPs.
Extending function approximation beyond linear models, \citet{ayoub2020model, wang2020reinforcement_eluder, ishfaq2021randomized} also prove regret bounds depending on Eluder dimension~\citet{russo2013eluder}.
There has been also some literature that aim to propose sample-efficient methods with more ``general'' function approximation. 
Yet, such claims may have been hindered by computational intractability \citet{krishnamurthy2016pac, jiang2017contextual, dann2018oracle} or having to rely on other stronger assumptions~\citet{du2019provably}, 
such that the resulting methods may turn out to be not as general or practical.

Despite the fact that there are a vast number of the existing works on RL with linear function approximation, there is very little work that extend beyond the linear model to other parametric models. 
To our best knowledge, \citet{wang2021glmRL} is the only existing work with generalized linear function approximation where the Bellman backup of any value function is assumed to be generalized linear function of feature mapping. \citet{wang2021glmRL} proposes a model-free algorithm under this assumption with $\tilde{\Ocal}(d^{3/2} H \sqrt{T})$ regret. 
In addition to the fact that their proposed method is model-free,
the significant difference between the problem setting of our work and that of \citet{wang2021glmRL} is that the transition probability in \citet{wang2021glmRL} is \textit{not} a generalized linear model, but rather generalized linear approximation is imposed directly on the value function update. 
Thus, the question of whether it is possible to design a provably efficient RL algorithm for an MDP with transition probability approximated by any generalized linear model including multinomial logistic model has still remained open.

\section{Preliminaries}

\subsection{Notations}
We denote by $[n]$ the set $\{1, 2, \ldots, n \}$ for a positive integer $n$. 
For a $d$-dimensional vector $\xb \in \RR^d$, we use $ \|\xb\|_2$ to denote the Euclidean norm of $\xb$.
The weighted $\ell_2$-norm associated with a positive definite matrix $\Ab$ is denoted by $\|\xb\|_\Ab := \sqrt{\xb^\top \Ab \xb}$. 
The minimum and maximum eigenvalues of a symmetric matrix $\Ab$ are written as $\lambda_{\min} (\Ab)$ and $\lambda_{\max} (\Ab)$ respectively.
The trace of a matrix $\Ab$ is $\tr(\Ab)$.
For two symmetric matrices $\Ab$ and $\Bb$ of the same dimensions, $\Ab \succcurlyeq \Bb$ means that $\Ab - \Bb$ is positive semi-definite.

\subsection{Problem Formulation}

We consider episodic Markov decision processes (MDPs) denoted by $\Mcal(\Scal, \Acal, H, P, r)$, 
where $\Scal$ is the state space, $\Acal$ is the action space, $H$ is the length of horizon, $P$ is the collection of transition probability distributions, and $r$ is a reward function.
Every episode starts at some initial state $s_1$ and ends after $H$ steps.
Then for every step $h \in [H]$ in an episode, the learning agent interacts with an environment defined by $\Mcal$, where the agent observes state $s_h \in \Scal$, selects an action $a_h \in \Acal$, and receives an immediate reward $r(s_h, a_h) \in [0,1]$.
And then, the next state $s_{h+1}$ is drawn from the transition probability distribution $P( \cdot \mid s_h, a_h)$ and repeats its interactions until the end of the episode, followed by a newly started episode.
A policy $\pi : \Scal \times [H] \rightarrow \Acal$ is a function that determines which action the agent takes in state $s_h$ at each step $h \in [H]$, $a_h \sim \pi(s_h, h)$.
Then, we define the value function of policy $\pi$, $V^{\pi}_h : \Scal \rightarrow \RR$ as the expected sum of rewards under the policy $\pi$ until the end of the episode when starting from $s_h = s$, i.e.,
\begin{equation*} \label{eq:value function}
    V^{\pi}_h(s) := \EE_\pi \left[ \sum_{h'=h}^{H} r \left(s_{h'}, \pi(s_{h'},h') \right) \mid s_{h} = s
    \right] \, .
\end{equation*}
We define the action-value function of policy $\pi$, $Q^\pi_h: \Scal \times \Acal \rightarrow \RR$ as the expected sum of rewards when following $\pi$ starting from step $h$ until the end of the episode after taking action~$a$ in state~$s$,
\begin{equation*}
    Q^\pi_h (s, a) := \EE_\pi \left[ \sum_{h'=h}^{H} r \left(s_{h'}, \pi(s_{h'},h') \right) \mid s_{h} = s, a_h = a \right] \, .
\end{equation*}

We define an optimal policy $\pi^*$ to be a policy that achieves the highest possible value at every state-step pair $(s,h) \in \Scal \times [H]$.
We denote by $V^*_h (s) = V^{\pi^*}_h (s)$ and $Q^*_h(s,a) = Q^{\pi^*}_h(s,a)$ as the the optimal value function and the optimal action-value function, respectively.
To make notation simpler, we denote $ P_h V_{h+1}(s,a):= \EE_{s' \sim P_h(\cdot \mid s,a)} [V_{h+1}(s')]$.
Recall that both $Q^\pi$ and $Q^*$ can be written as the result of the Bellman equations as
\begin{equation*}
    Q^\pi_h(s,a) = (r + P_h V^\pi_{h+1})(s,a) \, ,
    \quad
    Q^*_h(s,a) = (r + P_h V^*_{h+1})(s,a) \, ,
\end{equation*}
where $V^\pi_{H+1} (s) = V^*_{H+1} (s) = 0$ and $V^*_h(s) = \max_{a \in \Acal} Q^*_h(s,a)$ for all $s \in \Scal$.

The goal of the agent is to maximize the sum of future rewards, i.e., to find an optimal policy, through the repeated interactions with the environment for $K$ episodes.
Let policy $\pi = \{\pi_k\}_{k=1}^K$ be a collection of policies over $K$ episodes, where $\pi_k$ is the policy of the agent at $k$-th episode.
Then, the cumulative regret of $\pi$ over $K$ episodes is defined as
\begin{equation*}
    \Regret(K) := \sum_{k=1}^{K} (V^*_1 - V^{\pi_k}_1) (s_{k,1}) \, ,
\end{equation*}
where $s_{k,1}$ is the initial state in the $k$-th episode.
Therefore, maximizing the cumulative rewards of policy $\pi$ over $K$ episodes is equivalent to minimizing the cumulative regret $\Regret(K)$.

In this paper, we make a structural assumption for the MDP $\Mcal(\Scal, \Acal, H, P, r)$ where the transition probability kernel is given by an MNL model. Before we formally introduce the MNL function approximation,
we first introduce the following definition.

\begin{definition}[Reachable states]
    For each $(s,a) \in \Scal \times \Acal$, we define the reachable states of $(s,a)$ to be the set of all states which can be reached by taking action $a$ in state $s$ within a single transition, $\Scal_{s,a} := \{ s' \in \Scal : P(s' \mid s, a) \ne 0 \}$.
    Also, we denote $\Ucal := \max_{(s,a) \in \Scal \times \Acal} |\Scal_{s,a}|$
    to be the maximum size of reachable states.
\end{definition}

It is possible that even when the size of the state space $|\Scal|$ is very large, $\Ucal$ is still small. 
For example, consider the RiverSwim problem (Figure~\ref{fig:riverswim env}) with exponentially large state space. However, $\Ucal$ would be still 3, regardless of the state space size.    

\begin{assumption}[MNL transition models]\label{assum:mnl model}
    For given feature map $\varphib: \Scal \times \Acal \times \Scal \rightarrow \RR^d$, we assume that the probability of state transition to $s' \in \Scal_{s,a}$ when an action $a$ is taken at a state $s$ is given by,
    \begin{equation} \label{eq_transition model assumption}
        P(s' \mid s, a) = \frac{\exp(\varphib(s,a,s')^\top \thetab^*)}{\sum_{\tilde{s} \in \Scal_{s,a}} \exp(\varphib(s,a,\tilde{s})^\top \thetab^*)}
    \end{equation} 
    where $\thetab^* \in \RR^{d}$ is an unknown transition core parameter.
\end{assumption}

\begin{remark} \label{remark_zero feature}
    Without loss of generality, we assume that for all $(s,a) \in \Scal \times \Acal$, $\exists \dot{s} \in \Scal_{s,a}$ such that $\varphib(s,a,s') = \zero_d$.
    This is because suppose that a feature map $\phib(s,a,s') \in \RR^d$ is given and the transition probability is given by~\eqref{eq_transition model assumption} using the feature map $\phib$, i.e., $P(s' \mid s,a) = \frac{ \exp(\phib(s,a,s')^\top \thetab^*)}{\sum_{\tilde{s} \in \Scal_{s,a}} \exp(\phib(s,a,\tilde{s})^\top \thetab^*) }$.
    Then, by dividing the denominator and numerator by $\exp(\phib(s,a,\dot{s})^\top \thetab^*)$, and by defining $\varphib(s,a,\tilde{s}) := \phib(s,a,\tilde{s}) - \phib(s,a,\dot{s})$ we have a well-defined feature map $\varphib$ which induces a consistent transition probability defined by $\phib$, but $\varphib(s,a,\dot{s})=\zero_d$.
\end{remark}

In order to focus on the main challenge of model-based RL, we assume, without loss of generality,  that the reward function~$r$ is known for the sake of simplicity.\footnote{Note that this is without loss of generality since learning $r$ is much easier than learning $P$.}
This assumption on $r$ is standard in the model-based RL literature~\citet{yang2019sample, yang2020reinforcement, zhou2021nearly}.

\subsection{Linear Transition Model vs. Multinomial Logistic Transition Model}\label{sec:linear_vs_MNL}

In this section, we show how the linear model assumption is restrictive for the transition model of MDPs.
To our knowledge, this is the first rigorous discussion on the limitation of the linear transition model.
We first show that for an arbitrary set of features, a linear transition model (including bilinear and low-rank linear MDPs) cannot induce a proper probability distribution over next states.

\begin{proposition}\label{prop:arbitrary linear transition model}
    For an arbitrary set of features of state and actions of an MDP, there exist no linear transition model that can induce a proper probability distribution over next states.
\end{proposition}

Therefore, the linear model cannot be a proper choice of transition model in general. 
The restrictive linearity assumption on the transition model also affects the regret analysis of algorithms that are proposed under that assumption.
As an example, we show that one of the recently proposed model-based algorithms using the linear function approximation cannot avoid the dependence on the size of the state space $|\mathcal{S}|$.
~\citet{yang2020reinforcement} assumes the transition probability kernel is given by the bilinear interaction of the state-action feature $\phi$, the next state feature $\psi$, and the unknown transition core matrix $\Mb^*$, i.e., $P(s' | s, a) = \phib(s,a)^\top \Mb^* \psib(s')$. 
Before we show the suboptimal dependence, one can see that it is difficult to ensure that the estimated transition probability satisfies one of the fundamental probability properties, $\sum_{s'} \hat{P}(s' \mid s,a)=1$ based on Proposition~\ref{prop:arbitrary linear transition model}.
In the following proposition, we show that the regret of the proposed algorithm in~\citet{yang2020reinforcement}  actually depends linearly on the size of the state space despite the use of function approximation.

\begin{proposition}\label{prop:regret of matrixRL}
    The \text{MatrixRL} algorithm~\cite{yang2020reinforcement} based on the linear model has the regret of $\tilde{\Ocal}(|\mathcal{S}| H^2 d^{3/2} \sqrt{T})$ where $d$ is the dimension of the underlying parameter.
\end{proposition}

Hence, the bilinear model-based method cannot scale well with the large state space.
On the other hand, the MNL model defined in~\eqref{eq_transition model assumption} can naturally captures the categorical distribution for any feature representation of states and actions and for any parameter choice.
This is because, due to the normalization term of the MNL model, i.e., $\sum_{\tilde{s}} \exp(\varphib(s,a, \tilde{s})^\top \thetab)$ --- even if any estimated parameter for $\thetab^*$ is used to estimate the transition probability, the sum of the transition probabilities is always 1. This holds not only for the true transition model but also for the estimated model.
Hence, the MNL function approximation offers a more sensible model of the transition probability.

\section{Algorithm and Main Results}

\subsection{{\AlgName}  Algorithm}
\begin{algorithm}[!] 
    \caption{Upper Confidence Model-based RL for MNL Transition Model ($\AlgName$)} \label{alg_main algorithm}
    \begin{algorithmic}[1]
        \STATE \textbf{Inputs:} An episodic MDP $\Mcal$, Feature map $\varphib:\Scal \times \Acal \times \Scal \rightarrow \RR^{d}$, Total number of episodes~$K$, Regularization parameter $\{\lambda_k\}_{k=1}^K$, Confidence radius $\{ \beta_k \}_{k=1}^{K}$.
        \STATE \textbf{Initialize:} $\Ab_1 = \lambda_1 \Ib_d$, $\hat{\thetab}_1 = \zero_d$
        \FOR{episode $k=1,2, \cdots, K$}
            \STATE Set $\{ \hat{Q}_{k,h} \}_{h=1}^{H}$ as described in (\ref{eq:UC Q-function closed}) using $\hat{\thetab}_k$, $\beta_k$
            \FOR{horizon $h=1,2, \cdots, H$}
                \STATE Select an action $a_{k,h} = \argmax_{a \in \Acal} \hat{Q}_{k,h}(s_{k,h}, a)$ and observe $s_{k,h+1}$, $y_{k,h}$
            \ENDFOR
            \STATE Update $\Ab_{k+1} = \Ab_{k} + \sum_{h \le H} \sum_{s' \in \Scal_{k,h}} \varphib_{k,h,s'} \varphib_{k,h,s'}^\top$
            \STATE Compute $\hat{\thetab}_{k+1} = \argmin_{\thetab \in \RR^d}  \ell_{k+1} (\thetab)$
        \ENDFOR
    \end{algorithmic}
\end{algorithm}

\paragraph{Estimation of Transition Core}
Each transition sampled from the transition model provides information to the agent that updates the estimate for the transition core based on observed samples. 
For the sake of simple exposition, we assume discrete state space so that for all $k \in [K], h \in [H]$, we define the transition response variable $y_{k,h} = ( y_{k,h}^{s'} )_{s' \in \Scal_{k,h}}$ as $y_{k,h}^{s'} = \ind(s_{k,h+1} = s')$ for $s' \in \Scal_{s_{k,h}, a_{k,h}} =: \Scal_{k,h}$.
Then the transition response variable $y_{k,h}$ is a sample from the following multinomial distribution:
\begin{equation*}
    y_{k,h} \sim \text{multinomial} \big(1, [p_{k,h}(s', \thetab^*)]_{s' \in \Scal_{k,h}} \big) \, ,    
\end{equation*}
where the parameter $1$ indicates that $y_{k,h}$ is a \textit{single-trial} sample, and each probability $p_{k,h}(s', \thetab^*)$ is defined as
\begin{equation*}
    p_{k,h}(s', \thetab^*):= \frac{\exp(\varphib(s_{k,h}, a_{k,h}, s')^\top \thetab^*)} {\sum_{\tilde{s} \in \Scal_{k,h}}\exp(\varphib(s_{k,h}, a_{k,h}, \tilde{s})^\top \thetab^*)} \, .
\end{equation*}

Also, we define noise $\epsilon_{k,h}^{s'} := y_{k,h}^{s'} - p_{k,h}(s', \thetab^*)$.
Note that since $\epsilon_{k,h}^{s'}$ is bounded in $[-1,1]$, $\epsilon_{k,h}^{s'}$ is $1$-sub-Gaussian.

We estimate the unknown transition core $\thetab^*$ using the regularized \textit{maximum likelihood estimation} (MLE) for the MNL model. 
Based on the transition response variable $y_{k,h}$, the negative log-likelihood function of $\thetab$ with regularization parameter $\lambda_k > 0$ is then given by
\begin{equation} \label{eq:regularized negative log-likelihood}
    \ell_{k}(\thetab) = - \sum_{\substack{k'<k \\ h \le H}} \sum_{s' \in \Scal_{k',h}}\! y_{k', h}^{s'} \log p_{k', h}(s', \thetab)
    + \frac{\lambda_k}{2} \| \thetab \|_2^2\, .    
\end{equation}

Then the estimated transition core $\hat{\thetab}_k$ for $\thetab^*$ is given as follows:
\begin{equation}\label{eq_mle theta}
    \hat{\thetab}_k = \argmin_{\thetab \in \RR^d} \ell_k (\thetab) \,.
\end{equation}

\paragraph{Model-Based Upper Confidence}
To balance the exploration-exploitation trade-off, we construct an upper confidence action-value function which is greater than the optimal action-value function with high probability.
The upper confidence bounds (UCB) approaches are widely used due to their effectiveness in balancing the exploration and exploitation trade-off not only in bandit problems~\cite{auer2002using, auer2002finite, dani2008stochastic, filippi2010parametric, abbasi2011improved, chu2011contextual, li2017provably, zhou2020neural} but also in RL with function approximation~\cite{wang2021glmRL, jin2020provably, ayoub2020model, jia2020model}.

At the $k$-th episode, the confidence set $\Ccal_k$ for $\thetab^*$ is constructed based on the feature vectors that have been collected so far.
For previous episode $k' < k$ and the horizon step $h \le H$, we denote the associated features by $\varphib(s_{k',h}, a_{k',h}, s')=: \varphib_{k', h, s'}$ for $s' \in \Scal_{k', h}$.
Then from all previous episodes, we have $\{ \varphib_{k',h, s'}: s' \in \Scal_{k',h}, k' < k, h \le H \}$ and the observed transition responses of $\{ y_{k',h} \}_{k' < k, h \le H}$.
Let $\hat{\thetab}_k$ be the estimate of the unknown transition core $\thetab^*$ at the beginning of $k$-th episode, and suppose that we are guaranteed that $\thetab^*$ lies within the confidence set $\Ccal_k$ centered at $\hat{\thetab}_k$ with radius $\beta_k > 0$ with high probability.
Then for $(s,a) \in \Scal \times \Acal$ and for all $h \in [H]$, we construct the optimistic value function as follows:
\begin{equation} \label{eq:UC Q-function}
    \begin{aligned}
        & \hat{Q}_{k, H+1}(s,a) := 0 \, ,
        \\
        & \hat{Q}_{k, h}(s, a) := r(s,a) + \max_{\thetab \in \Ccal_k}  \sum_{s' \in \Scal_{s,a}} \frac{ \exp\!\big(\varphib(s,a,s')^\top \thetab\big) \hat{V}_{k,h+1}(s')}{\sum_{\tilde{s} \in \Scal_{s,a}} \exp\!\big(\varphib(s,a,\tilde{s})^\top \thetab \big)} \, ,                        
    \end{aligned}
\end{equation}
where $\hat{V}_{k,h}(s):= \min \big\{ \max_a \hat{Q}_{k,h}(s, a), H \big\}$.
Also, the confidence set $\Ccal_k$ for $\thetab^*$ is constructed as
\begin{equation*} 
    \Ccal_k := \{ \thetab \in \RR^{d}: \| \thetab - \hat{\thetab}_k \|_{\Ab_k} \le \beta_k (\delta) \} \, ,
\end{equation*}
where radius $\beta_k$ is specified later, and 
the Gram matrix $\Ab_k$ is given for some $\lambda_k >0$ by 
\begin{equation} \label{eq:gram matrix}
    \Ab_k = \lambda_k \Ib_d + \sum_{\substack{k'<k \\ h \le H}} \sum_{s' \in \Scal_{k',h}}\! \varphib_{k',h, s'} \varphib_{k',h, s'}^\top \, .
\end{equation}

As long as the true transition core $\thetab^* \in  \Ccal_k$ with high probability, the action-value estimates defined as~\eqref{eq:UC Q-function} are optimistic estimates of the actual $Q$ values.
Based on these optimistic action-values $\{ \hat{Q}_{k,h} \}_{h=1}^{H}$, 
in each $h \in [H]$ time step of the $k$-th episode, the agent selects the greedy action $a_{k,h} = \argmax_{a \in \Acal} \hat{Q}_{k,h} (s_{k,h}, a)$.
The full algorithm is summarized in Algorithm~\ref{alg_main algorithm}.

\paragraph{Closed-Form UCB}
When we construct the optimistic value function as described in~\eqref{eq:UC Q-function}, it is required to solve a maximization problem over a confidence set.
However, an explicit solution of this maximization problem is not necessary.
The algorithm only requires the estimated action-value function to be optimistic.
We can use a closed-form confidence bound instead of computing the maximal $\thetab$ over the confidence set.
Also, we can verify that the regret bound in  Theorem~\ref{thm_main theorem} still holds even when we replace~\eqref{eq:UC Q-function} with the following equation: for all $h \in [H]$, 
\begin{equation}
    \begin{split} \label{eq:UC Q-function closed}
        \hat{Q}_{k,h}(s,a) = r(s,a) +  \sum_{s' \in \Scal_{s,a}} \frac{\exp\!\big(\varphib(s,\!a,\!s')^\top \hat{\thetab}_k\big) \hat{V}_{k,h\!+\!1}(s')}{\sum_{\tilde{s} \in \Scal_{s,a}} \exp\!\big(\varphib(s,\!a,\! \tilde{s})^\top \hat{\thetab}_k\big)} + 2 H \beta_k \!\max_{s' \in \Scal_{s,a}} \!\!\| \varphib(s,\!a,\!s') \|_{\Ab_k^{-1}} \, . 
    \end{split}    
\end{equation}

\subsection{Regret Bound for {\AlgName} Algorithm}
In this section, we present the regret upper-bound of $\AlgName$.
We first start by introducing the standard regularity assumptions.
\begin{assumption}[Boundedness]\label{asm_feature norm}
    There exist $L_{\varphib}, L_{\thetab} > 0$ such that $\| \varphib(s,a,s') \|_2 \le L_{\varphib}$ for all $(s,a) \in \Scal \times \Acal$, $s' \in \Scal_{s,a}$, and $\|\thetab^*\|_2 \le L_{\thetab}$.
\end{assumption}
Note that this assumption is used to make the regret bounds scale-free for convenience and is in fact standard in the literature of RL with function approximation~\citet{jin2020provably, yang2020reinforcement, zanette2020frequentist}.

\begin{assumption}\label{asm_kappa}
    There exists $0 < \kappa < 1$ such that for all $(s,a) \in \Scal \times \Acal$ and $s', s'' \in \Scal_{s,a}$ and for all $k \in [K], h \in [H]$, $\inf_{\thetab \in \RR^{d}} p_{k,h}(s', \thetab) p_{k,h}(s'', \thetab) \ge \kappa$.
\end{assumption}

Assumption~\ref{asm_kappa} is equivalent to a standard assumption in generalized linear contextual bandit literature~\citet{filippi2010parametric, desir2014near, li2017provably, oh2019thompson, kveton2020randomized, russac2020algorithms, oh2021multinomial} to guarantee the Fisher information matrix is non-singular and is modified to suit our setting. 

Under these regularity conditions, we state the regret bound for Algorithm~\ref{alg_main algorithm}.
\begin{theorem}[Regret bound of $\AlgName$]\label{thm_main theorem}
    Suppose that Assumptions~\ref{assum:mnl model}-\ref{asm_kappa} hold. 
    For $\delta \in (0,1)$, if we set the input parameters of Algorithm~\ref{alg_main algorithm} as
    \begin{align*}
        \lambda_k = \Ocal(L_{\varphib}^2 d \log (k H \Ucal L_{\varphib}^2)) \, ,
        \quad 
        \beta_k = \Ocal(\sqrt{d} \log (k H \Ucal L_{\varphib}^2/\delta)) \, ,
    \end{align*}
    then with probability at least $1-\delta$, the cumulative regret of the $\AlgName$ policy $\pi$ is upper-bounded by
    \begin{equation*}
        \Regret(K) = \tilde{\Ocal} \left( \kappa^{-1} d \sqrt{H^3 T} \right) \, ,
    \end{equation*}
    where $T = KH$ is the total number of time steps.
\end{theorem}

\paragraph{Discussion of Theorem~\ref{thm_main theorem}}

In terms of the key problem primitives, Theorem~\ref{thm_main theorem} states that $\AlgName$ achieves $\tilde{\Ocal} ( d \sqrt{H^3 T})$ regret. 
To our best knowledge, this is the first result to guarantee a regret bound for the MNL model of the transition probability kernel.
Among the existing model-based methods with function approximation, the most related method to ours is a bilinear matrix-based algorithm in~\citet{yang2020reinforcement}. \citet{yang2020reinforcement} shows $\tilde{\Ocal}(d^{3/2} H^2 \sqrt{T})$ regret under the assumption that the transition probability can be a linear model parametrized with an unknown transition core matrix. 
Hence, the regret bound in Theorem~\ref{thm_main theorem} is sharper in terms of dimensionality and the episode length. 
Furthermore,
as mentioned in Proposition~\ref{prop:arbitrary linear transition model}, the regret bound in~\citet{yang2020reinforcement} contains additional $|\mathcal{S}|$ dependence. Therefore, our regret bound shows an improved scalability over the method developed under a similar model.
On the other hand, for linear mixture MDPs \citet{jia2020model,ayoub2020model,zhou2021nearly}, the lower bound of $\Omega(d H \sqrt{T})$
has been proven in \citet{zhou2021nearly}. 
Hence, noting the total dimension of the unknown parameter, 
the dependence on dimensionality as well as dependence on total steps matches the corresponding dependence in the regret bound for linear MDP~\citet{zhou2021nearly}.
This provides a conjecture that the dependence on $d$ and $T$ in Theorem~\ref{thm_main theorem} is best possible although a precise lower bound in our problem setting has not yet been shown.

\subsection{Proof Sketch and Key Lemmas}
In this section, we provide the proof sketch of the regret bound in Theorem~\ref{thm_main theorem} and the key lemmas for the regret analysis.
In the following lemma, we show that the estimated transition core $\hat{\thetab}_k$ concentrates around the unknown transition core $\thetab^*$ with high probability.
\begin{lemma}[Concentration of the transition core]~\label{lemma_theta hat concentration} 
    Suppose that Assumptions~\ref{assum:mnl model}-\ref{asm_kappa} hold.
    For $\delta \in (0,1)$, with probability at least $1 - \delta$, we have for all $k \ge 1$,
    \begin{equation*}
        \thetab^* \in \Ccal_k := \left\{ \thetab \in \RR^{d}: \| \thetab - \hat{\thetab}_k \|_{\Ab_k} \le \beta_k (\delta) \right\} \, ,
    \end{equation*}
    where
    \begin{equation*}
        \beta_k(\delta) := \kappa^{-1} \left( \bigg( \frac{L_{\varphib}}{4} + L_{\thetab}  \bigg) \sqrt{\lambda_k}
         + \frac{4}{L_{\varphib}\sqrt{\lambda_k}} \left( d \log \frac{2}{L_{\varphib}^2} + \frac{d}{2} \log \left( 1 + \frac{k H \Ucal L_{\varphib}^2}{d \lambda_k} \right) - \log \delta \right) \right) \, .
    \end{equation*}
\end{lemma}
Then, we show that when our estimated parameter $\hat{\thetab}_k$ is concentrated around the transition core $\thetab^*$, the estimated upper confidence action-value function is deterministically greater than the true optimal action-value function. That is, the estimated $\hat{Q}_{k,h} (s,a)$ is optimistic.
\begin{lemma}[Optimism] \label{lemma_optimism of estimated q function}
    Suppose that Lemma~\ref{lemma_theta hat concentration} holds. 
    Then for all $(s, a) \in \Scal \times \Acal$ and $h \in [H]$, we have
    \begin{equation*}
        Q^*_h (s,a) \le \hat{Q}_{k,h} (s,a) \, .
    \end{equation*}
\end{lemma}
This optimism guarantee is crucial because it allows us to work with the estimated value function which is under our control rather than working with the unknown optimal value function.
Next, we show that the error induced by the value iteration per step is bounded.

\begin{lemma}[Concentration of the value function]\label{lemma:value f.t concentration}
    Suppose that Lemma~\ref{lemma_theta hat concentration} holds.
    For $\delta \in (0,1)$ and for any $h \in [H]$, we have
    \begin{align*}
        & \hat{Q}_{k,h} (s_{k,h}, a_{k,h}) - \left[ r(s_{k,h}, a_{k,h}) + P_h \hat{V}_{k, h+1} (s_{k,h}, a_{k,h}) \right] 
        \le 2 H \beta_k (\delta) \max_{s' \in \Scal_{k,h}} \| \varphib_{k,h,s'} \|_{\Ab_k^{-1}} \, .        
    \end{align*}
\end{lemma}
With these lemmas at hand, by summing per-episode regrets over all episodes and by the optimism of the estimated value function, the regret can be bounded by the sum of confidence bounds on the sample paths. 
All the detailed proof is deferred to Appendix~\ref{appx_proof of thm1}.

\section{Numerical Experiments}~\label{sec:numerical experiments}

In this section, we evaluate the performances of our proposed algorithm, $\AlgName$ in numerical experiments.
We perform our experiments on the RiverSwim~\citet{osband2013more} environment. 
\begin{figure}[H]
    \tikzstyle{every node}=[font=\small]
        \centering
        \resizebox{\columnwidth}{!}{
        \begin{tikzpicture}[->,>=stealth',shorten >=2pt, 
        line width=0.7 pt, node distance=1.6cm,
        scale=1, 
        transform shape, align=center, 
        state/.style={circle, minimum size=0.5cm, text width=5mm}]
    \node[state, draw] (one) {$s_1$};
    \node[state, draw] (two) [right of=one] {$s_2$};
    \node[state, draw=none] (dots) [right of=two] {$...$};
    
    \node[state, draw] (n-1) [right of=dots] {$s_{n-1}$};
    \node[state, draw] (n) [right of=n-1] {$s_{n}$};

    \path (one) edge [ loop above ] node {$0.4$} (one) ;
    \path (one) edge [ bend left ] node [above]{$0.6$} (two) ;
    \draw[->] (two.145) [bend left] to node[below]{$0.05$} (one.35);
    \path (two) edge [ loop above ] node {$0.6$} (two) ;
    \path (two) edge [ bend left ] node [above]{$0.35$} (dots) ;
    \draw[->] (dots.145) [bend left] to node[below]{$0.05$} (two.35); 
    \path (n-1) edge [ loop above ] node {$0.6$} (n-1) ;
    \path (dots) edge [ bend left ] node [above]{$0.35$} (n-1) ;
    \draw[->] (n-1.145) [bend left] to node[below]{$0.05$} (dots.35); 
    \path (n-1) edge [ bend left ] node [above]{$0.35$} (n) ;
    \draw[->] (n.145) [bend left] to node[below]{$0.4$} (n-1.35); 
    
    \path[densely dashed] (one) edge [ loop left ] node {$(1, r=\frac{5}{1000})$} (one) ;
    \draw[densely dashed, ->] (two.-100) [bend left] to node[below]{$1$} (one.-80);
    \draw[densely dashed, ->] (dots.-100) [bend left] to node[below]{$1$} (two.-80);
    \draw[densely dashed, ->] (n-1.-100) [bend left] to node[below]{$1$} (dots.-80);
    \draw[densely dashed, ->] (n.-100) [bend left] to node[below]{$1$} (n-1.-80);
    
    \path (n) edge [ loop right ] node {$(0.6, r=1)$} (n);    
    \end{tikzpicture}
    }
      \caption{The ``RiverSwim'' environment with $n$ states~\citet{osband2013more}}
    \label{fig:riverswim env}
\end{figure}
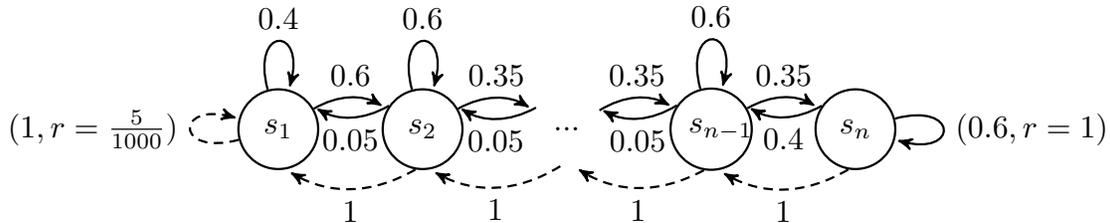

The RiverSwim environment is considered to be a challenging problem setting where na\"ive dithering approaches, such as the $\epsilon$-greedy policy, are known to have poor performance and require efficient exploration.
The RiverSwim in Figure~\ref{fig:riverswim env} consists of $n$ states (i.e., $|\Scal| = n$) lined up in a chain, with the number on each of the edges representing the transition probability.
Starting in the leftmost state $s_1$, the agent can choose to swim to the left --- whose outcomes are represented by the dashed lines --- and collect a small reward. i.e., $r(s_1,\texttt{left})=0.05$. Or, the agent can choose to swim to the right --- whose outcomes are represented by the \textit{solid} lines --- in each succeeding state. 
The agent's goal is to maximize its return by attempting to reach the rightmost state $s_n$ where a large reward $r(s_n,\texttt{right})=1$ can be obtained by swimming right.

Since the objective of this experiment is to see how efficiently our algorithm explores compared to other provably efficient RL algorithms with function approximation, we choose both model-based algorithms, \texttt{UC-MatrixRL}~\cite{yang2020reinforcement} and \texttt{UCRL-VTR}~\cite{ayoub2020model} and model-free algorithms, \texttt{LSVI-UCB}~\cite{jin2020provably} and \texttt{LSVI-PHE}~\cite{ishfaq2021randomized} for comparisons.
\begin{figure}[h]
    \centering
    \begin{subfigure}[b]{0.45\textwidth}
        \includegraphics[width=\textwidth]{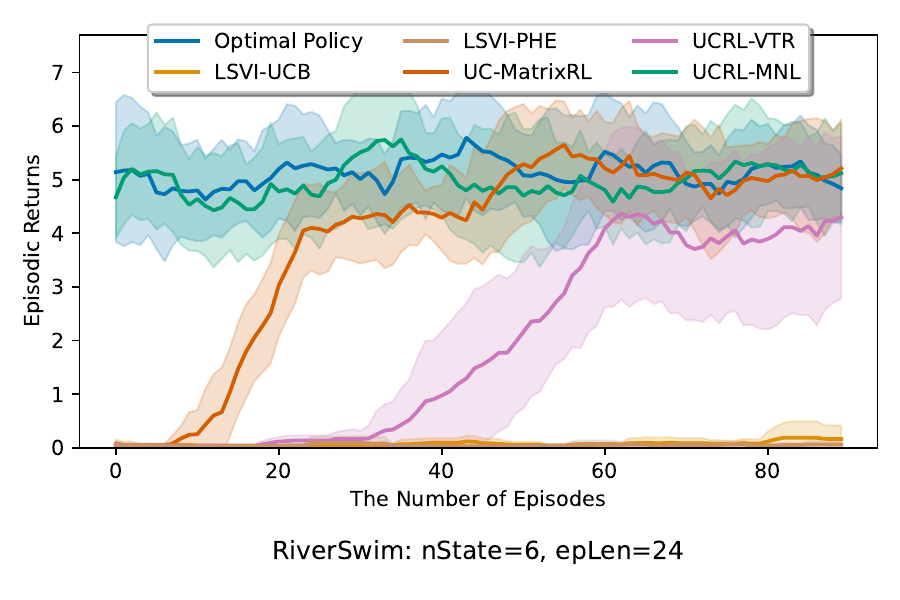}
    \end{subfigure}
    \begin{subfigure}[b]{0.45\textwidth}
        \includegraphics[width=\textwidth]{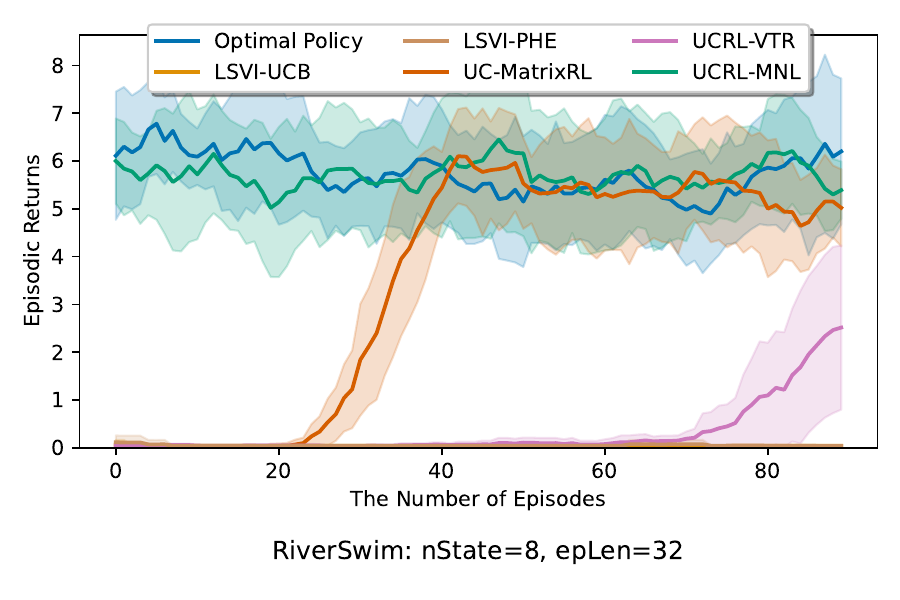}
    \end{subfigure}
    \begin{subfigure}[b]{0.45\textwidth}
        \includegraphics[width=\textwidth]{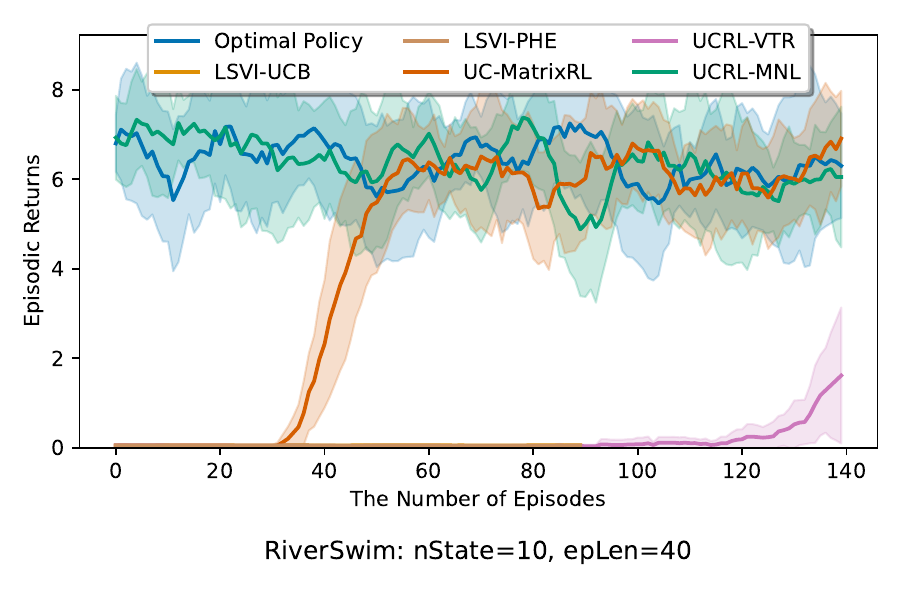}
    \end{subfigure}
    \begin{subfigure}[b]{0.45\textwidth}
        \includegraphics[width=\textwidth]{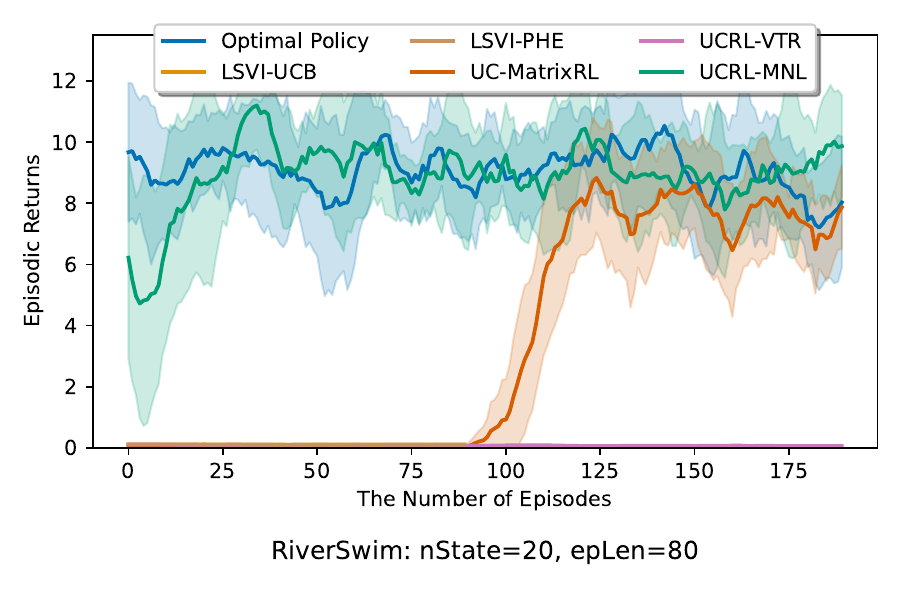}
    \end{subfigure}
    \caption[ The average and standard deviation of critical parameters ]
        {Episodic returns over 10 independent runs under the different RiverSwim environments} 
        \label{fig:RiverSwim}
\end{figure}

We perform a total of four experiments while increasing the number of states for RiverSwim.
To set the hyper-parameters for each algorithm, we performed a grid search over certain ranges. 
In each experiment, we evaluated the algorithms on 10 independent instances to report the average performance.
First, Figure~\ref{fig:RiverSwim} shows the episodic return of each algorithm over 10 independent runs.
When the number of states is small (e.g., $|\Scal| = 6$), it can be seen that not only our algorithm but also other model-based algorithms learn the optimal policy relatively well. However, our algorithm $\AlgName$ clearly outperforms the existing algorithms.
As the number of states increases (e.g., $|\Scal| = 20$),
we observe that our algorithm reaches the optimal values remarkably quickly compared to the other algorithms, outperforming the existing algorithms by significant margins.

Table~\ref{tab:cumulative reward} shows the average cumulative reward over the episodes of each algorithm for 10 independent runs.
The proposed algorithm has an average cumulative reward similar to that of the optimal policy across all problem settings.
The results of these experiments provide the evidence for the practicality of our proposed model and proposed algorithm.
For more detail, please refer to Appendix~\ref{appx:experiment details}.
\begin{table}[t]
    \resizebox{\columnwidth}{!}{
    \begin{tabular}{lrrrr}
         \toprule
          &  \multicolumn{4}{c}{RiverSwim Environment} \\
          \cmidrule(r){2-5}
         Methods  & $|\Scal|=6, H=24$ & $|\Scal|=8, H=32$ & $|\Scal|=10, H=40$ & $|\Scal|=20, H=80$ \\
         \midrule
         Optimal Policy  & 513.70 $\pm$ 39.67 & 576.50 $\pm$ 43.65 & 959.70 $\pm$ 57.97 & 1816.50 $\pm$ 79.89\\
        \midrule
         $\AlgName$ & 492.64 $\pm$ 32.23 & 562.83 $\pm$ 44.72 & 948.64 $\pm$ 74.47 & 1777.11 $\pm$ 99.23\\
         \texttt{UC-MatrixRL} & 372.47 $\pm$ 18.04 & 334.13 $\pm$ 21.74 & 632.62 $\pm$ 25.22 & 682.15 $\pm$ 64.41 \\
         \texttt{UCRL-VTR} & 191.72 $\pm$ 68.47 & 41.50 $\pm$ 27.81 & 28.59 $\pm$ 19.61 & 15.76 $\pm$ 0.848 \\         
         \texttt{LSVI-UCB} & 7.52 $\pm$ 5.284  & 4.42 $\pm$ 2.053 & 5.54 $\pm$ 1.123 & 10.97 $\pm$ 1.813 \\
         \texttt{LSVI-PHE} & 5.15 $\pm$ 1.506  & 5.41 $\pm$ 2.081  & 5.23 $\pm$ 0.097 & 10.17 $\pm$ 0.139\\
         \bottomrule
    \end{tabular}
    }
    \caption{Average returns over 10 independent runs in RiverSwim environments}
    \label{tab:cumulative reward}
\end{table}
%
\section{Conclusion}

In this work, we propose a new MNL transition model for RL with function approximation overcoming the limitation of the linear function approximation and a new algorithm, $\AlgName$.
We establish the theoretical performance bound on the proposed algorithm's regret. 
To the best of our knowledge, this is not only the first model-based RL algorithm with multinomial logistic function approximation, but also our algorithm enjoys the provable efficiency as well as superior practical performances.

\section*{Acknowledgments}

This work was supported by the National Research Foundation of Korea(NRF) grant funded by the Korea government(MSIT) (No. 2022R1C1C1006859) and by Creative-Pioneering Researchers Program through Seoul National University.

\bibliographystyle{plainnat}
\bibliography{references}

\clearpage
\appendix
\onecolumn

\renewcommand{\contentsname}{Contents of Appendix}
\addtocontents{toc}{\protect\setcounter{tocdepth}{2}}
{
  \hypersetup{hidelinks}
  \tableofcontents
}

\section{Further Discussion on the Limitations of the Linear Transition Model}\label{appx_constraint of linear model}
\subsection{Proof of Proposition~\ref{prop:arbitrary linear transition model}}

We provide counter-examples as a proof of Proposition~\ref{prop:arbitrary linear transition model} both for the bilinear transition case~\cite{yang2020reinforcement} and for the low-rank MDP case~\cite{jin2020provably}.

\textbf{Case 1) Bilinear transition \citet{yang2020reinforcement}.}
In the bilinear transition model, the transition probability kernel is given by 
\begin{equation*}
    \PP (s' \mid s, a) = \phib(s,a)^\top \Mb^* \psib(s')
\end{equation*}
where $\phib:\Scal \times \Acal \rightarrow \RR^d$, $\psib: \Scal \rightarrow \RR^{d'}$ are known feature maps and $\Mb^*$ is an unknown transition core matrix. 
Let an MDP with $|\Scal| = 2$ and $| \Acal| = 2 $ be given. Also, let us given the features of state-action $\phib(s,a) \in \RR^2$ and next state features $\psi(s') \in \RR$ as follows:
\begin{align*}
    & \Phi^\top := [\phib(s_1, a_1)^\top, \phib(s_1, a_2)^\top, \phib(s_2, a_1)^\top, \phib(s_2,a_2)^\top]^\top =
    \begin{bmatrix}
        1 & 1 \\
        1 & 2 \\
        3 & 1 \\
        2 & 3
    \end{bmatrix}
    \\
    & \Psi := [\psi(s_1), \psi(s_2)] = [1, -2] \, .
\end{align*}
Then for the true transition core matrix $\Mb^* \in \RR^{2\times1}$, the element-wise sum across each row of $\Phi^\top \Mb^* \Psi$ should be 1, because it is the transition matrix for the given MDP.
On the other hand, if let $\Mb^* = [x, y]^\top$, then, we have
\begin{equation*}
    \Phi^\top \Mb^* \Psi
    = 
    \begin{bmatrix}
        \PP(s_1 \mid s_1, a_1) & \PP(s_2 \mid s_1, a_1) \\
        \PP(s_1 \mid s_1, a_2) & \PP(s_2 \mid s_1, a_2) \\
        \PP(s_1 \mid s_2, a_1) & \PP(s_2 \mid s_2, a_1) \\
        \PP(s_1 \mid s_2, a_2) & \PP(s_2 \mid s_2, a_2) \\
    \end{bmatrix}
    =
    \begin{bmatrix}
        x+y & -2(x+y) \\
        x+2y & -2(x+2y) \\
        3x+y & -2(3x+y) \\
        2x+3y & -2(2x+3y) \\
    \end{bmatrix}\, .
\end{equation*}
Then by the property of the transition probability kernel, we have the following 4 equations:
\begin{equation*}
    -x -y = 1 \, , -x -2y = 1 \, , -3x -y = 1\, , -2x -3y = 1 \, .
\end{equation*}
However, this system of linear equations is inconsistent since there are more equations to be satisfied than the number of unknown variables. 
Hence, in general, no bilinear transition model can induce a proper probability distribution over next states for an arbitrary set of state and action features.

\textbf{Case 2) Low-rank MDP \citet{yang2019sample, jin2020provably, zanette2020frequentist}}
In the low-rank MDP, the transition probability is given by
\begin{equation*}
    \PP( \cdot \mid s,a) = \langle \phib(s,a), \boldsymbol{\mu}^* \rangle
\end{equation*}
where $\phib: \Scal \times \Acal \rightarrow \RR^d$ is a known feature map and $\boldsymbol{\mu}^* \in \RR^{d \times |\Scal|}$ is an unknown (signed) measure over $\Scal$. 
Let a low-rank MDP with $|\Scal|=2$ and $|\Acal|=2$ be given. Suppose that the features of state-action $\phi(s,a) \in \RR^3$ are given as follows:
\begin{align*}
    & \Phi^\top := [\phib(s_1, a_1)^\top, \phib(s_1, a_2)^\top, \phib(s_2, a_1)^\top, \phib(s_2,a_2)^\top]^\top =
    \begin{bmatrix}
        1 & 1 & 1 \\
        1 & 2 & 0 \\
        3 & 1 & 4 \\
        2 & 3 & 7
    \end{bmatrix}
\end{align*}
Then similarly with the Case 1, the element-wise sum across each row of $\Phi^\top \boldsymbol{\mu}^*$ should be 1.
If we let $\boldsymbol{\mu}^*$ as follows:
\begin{equation*}
    \boldsymbol{\mu}^* =
    \begin{bmatrix}
        x_{11} & x_{12} \\
        x_{21} & x_{22} \\
        x_{31} & x_{32} \\
    \end{bmatrix} \, ,
\end{equation*}
then, we have
\begin{equation*}
    \Phi^\top \boldsymbol{\mu}^*
    = 
    \begin{bmatrix}
        \PP(s_1 \mid s_1, a_1) & \PP(s_2 \mid s_1, a_1) \\
        \PP(s_1 \mid s_1, a_2) & \PP(s_2 \mid s_1, a_2) \\
        \PP(s_1 \mid s_2, a_1) & \PP(s_2 \mid s_2, a_1) \\
        \PP(s_1 \mid s_2, a_2) & \PP(s_2 \mid s_2, a_2) \\
    \end{bmatrix}
    =
    \begin{bmatrix}
        x_{11} + x_{21} + x_{31} & x_{12}+x_{22} + x_{32}\\
        x_{11} + 2x_{21} & x_{12} + 2 x_{22} \\
        3x_{11} + x_{21} + 4x_{31} & 3x_{12} + x_{22} + 4x_{32} \\
        2x_{11} + 3x_{21} + 7x_{31} & 2x_{12} + 3 x_{22} + 7 x_{32} \\
    \end{bmatrix}
    \, .
\end{equation*}
Then by the property of the transition probability kernel, we have the following 4 equations:
\begin{align*}
    & x_{11} + x_{12} + x_{21} + x_{22} + x_{31} + x_{32} = 1 \, ,
    \\
    & x_{11} + x_{12} + 2 x_{21} + 2 x_{22} = 1 \,,
    \\
    & 3 x_{11} + 3 x_{12} + x_{21} + x_{22} + 4 x_{31} + 4 x_{32} = 1 \, ,
    \\
    & 2 x_{11} + 2x_{12} + 3 x_{21} + 3x_{22} + 7 x_{31} + 7 x_{32} = 1 \, .
\end{align*}
On the other hand, if we put 
\begin{equation*}
    \Qb = 
    \begin{bmatrix}
        1 & 1 & 1 & 1 & 1 & 1 \\
        1 & 1 & 2 & 2 & 0 & 0 \\
        3 & 3 & 1 & 1 & 4 & 4 \\
        2 & 2 & 3 & 3 & 7 & 7 \\
    \end{bmatrix}
    \text{ and }
    \bb =     
    \begin{bmatrix}
        1 \\
        1 \\
        1 \\
        1 \\
    \end{bmatrix} \,
\end{equation*}
then we have $\rank[\Qb \mid \bb] = 4$ and $\rank \Qb = 3$. 
Since the rank of the augmented matrix $[\Qb \mid \bb]$ is greater than the rank of the coefficient matrix $\Qb$,
by the Rouch\'{e}-Capelli theorem~\cite{shafarevich2012linear}, the system of linear equations is inconsistent.
Hence, in general, no transition model for a low-rank MDP can induce a proper probability distribution over next states for an arbitrary set of state and action features.

\subsection{Proof of Proposition \ref{prop:regret of matrixRL}}
For the analysis of regret bound in~\citet{yang2020reinforcement}, the authors impose regularity assumptions on features. 
In particular,~\citet{yang2020reinforcement} define $\Psi = [ \psib(s_1), \ldots, \psib(s_{|\Scal|})]^\top \in \RR^{|\Scal| \times d'}$ as the matrix of all features $\psib$ of next states and assume $\Psi$ can satisfy $\| \Psi^\top v \|_{\infty} \le C_\psi \|\vb\|_\infty$ for some universal constant $C_\psi > 0$ and for any $\vb \in \RR^{|\Scal|}$.
However, each component of $\Psi^\top v$ consists of the sum of a total of $|\Scal|$ elements, thus the upper bound of $\| \Psi^\top v \|_\infty$ should depend on $| \Scal |$, i.e., $\| \Psi^\top v \|_{\infty} \le C_\psi |\Scal| \|v\|_\infty$.
Hence, the regret bound in~\citet{yang2020reinforcement} would increase by a factor of $|\Scal|$.

\section{Proof of Theorem~\ref{thm_main theorem}} \label{appx_proof of thm1}
In this section, after presenting all the necessary analyses of the lemmas required for proving Theorem~\ref{thm_main theorem}, we provide the proof of Theorem~\ref{thm_main theorem} at the end.

\subsection{Proof of Lemma~\ref{lemma_theta hat concentration}}
As mentioned by~\citet{li2024provably}, there is a technical issue related to Lemma 1 in the previous version ~\cite{hwang2023model}.
Specifically, when constructing the confidence set for the true transition core $\thetab^*$, the self-normalized bound for vector-valued martingales (Theorem 1 in~\citet{abbasi2011improved}) cannot be applied due to the correlation of the noise variables $\{ \epsilon^{s'}_{k,h} \}_{s' \in \Scal_{k,h}}$.
Instead, we construct the confidence set for $\thetab^*$ based on a new Bernstein-like tail inequality for self-normalized vectorial martingales with correlated noise~\cite{goyal2021dynamic, li2024improved}.
For completeness, we have revised the proof of Lemma~\ref{lemma_theta hat concentration}.

\begin{proof}[Proof of Lemma~\ref{lemma_theta hat concentration}]
    Let us define the function $\Gb_k(\thetab)$ as the difference in the gradients of the regularized negative log-likelihood in~\eqref{eq:regularized negative log-likelihood}:
    \begin{align*}
        \Gb_k(\thetab) & := \nabla \ell_k (\thetab) - \nabla \ell_k (\thetab^*)
        \\
        & = \sum_{\substack{k' < k, \\ h \le H}} \sum_{s' \in \Scal_{k',h}} [p_{k',h}(s', \thetab) - p_{k',h}(s', \thetab^*)] \varphib_{k', h, s'}
            + \lambda_k (\thetab - \thetab^*) \, .
    \end{align*}
    Then, it follows that 
    \begin{align*}
        \Gb_k(\hat{\thetab}_k) 
        & = \sum_{\substack{k' < k, \\ h \le H}} \sum_{s' \in \Scal_{k',h}} [p_{k',h}(s', \hat{\thetab}_k) - p_{k', h}(s', \thetab^*)] \varphib_{k', h, s'}
        + \lambda_k (\hat{\thetab}_k - \thetab^*)
        \\
        & = \sum_{\substack{k' < k, \\ h \le H}} \! \sum_{s' \in \Scal_{k',h}} \!
        \!\! [p_{k', h}(s', \hat{\thetab}_k) - y_{k',h}^{s'}] \varphib_{k', h, s'} + \lambda_k \hat{\thetab}_k
            \\
            & \quad + \sum_{\substack{k' < k, \\ h \le H}} \! \sum_{s' \in \Scal_{k',h}} \!\!\! [y_{k',h}^{s'} - p_{k', h}(s', \thetab^*)] \varphib_{k', h, s'}
        - \lambda_k \thetab^*
        \\
        & = \zero_d + \sum_{\substack{k' < k, \\ h \le H}} \sum_{s' \in \Scal_{k',h}} \epsilon_{k',h}^{s'} \varphib_{k',h, s'} - \lambda_k \thetab^*
    \end{align*}
    where $\epsilon_{k',h}^{s'} := y_{k',h}^{s'} - p_{k', h}(s', \thetab^*)$ and the second equality holds since $\hat{\thetab}_k$ is the solution of the following equation:
    \begin{equation*}
        \nabla \ell_k (\hat{\thetab}_k) = \sum_{\substack{k' < k, \\ h \le H}} \sum_{s' \in \Scal_{k',h}} [p_{k',h}(s', \hat{\thetab}_k) - y_{k',h}^{s'}] \varphib_{k', h, s'} + \lambda_k \hat{\thetab}_k = \zero_d \, .
    \end{equation*}
    On the other hand, by Remark~\ref{remark_zero feature}, if we denote a state in the set of reachable states $\Scal_{k',h}$ satisfying $\varphib(s, a, s') = \zero$ as $\dot{s}_{k',h}$, then we have
    \begin{align*}
        p_{k',h} (s', \thetab)
        & = \frac{\exp(\varphib_{k',h,s'}^\top \thetab)} { \sum_{s'' \in \Scal_{k',h}} \exp(\varphib_{k',h,s''}^\top \thetab)}
        \\
        & = \frac{\exp ( \varphib_{k', h, s'}^\top \thetab )}{ 1 + \sum_{s'' \in \Scal_{k',h} \backslash \{ \dot{s}_{k',h} \}} \exp ( \varphib_{k',h,s''}^\top \thetab)} \, .
    \end{align*}    
    Notice that for $j \in [d]$, we have
    \begin{align*}
        \frac{\partial{p_{k',h}(s', \thetab)}}{\partial{\thetab_j}}
        = p_{k',h}(s', \thetab) \varphib_{k', h, s'}^{(j)}
        - p_{k', h}(s', \thetab) \sum_{s'' \in \Scal_{k',h} \backslash \{ \dot{s}_{k',h} \}} p_{k', h}(s'', \thetab)  \varphib_{k',h,s''}^{(j)} \, ,
    \end{align*}
    which implies,
    \begin{align*}
        \nabla_{\thetab} p_{k',h} (s', \thetab) = p_{k',h} (s', \thetab) \varphib_{k',h, s'} - p_{k',h}(s', \thetab)
        \sum_{s'' \in \Scal_{k',h} \backslash \{ \dot{s}_{k',h} \} } p_{k',h}(s'', \thetab) \varphib_{k',h, s''} \, .
    \end{align*}
    Then for any $\thetab_1, \thetab_2 \in \RR^{d}$, the mean value theorem implies that there exists $\bar{\thetab} = \nu \thetab_1 + (1-\nu) \thetab_2$ with some $\nu \in (0,1)$ such that
    \begin{align*}
        \Gb_k(\thetab_1) - \Gb_k(\thetab_2)
        & =  \sum_{\substack{k' < k, \\ h \le H}} \sum_{s' \in \Scal_{k',h}} [p_{k',h}(s', \thetab_1) - p_{k',h}(s', \thetab_2)] \varphib_{k', h, s'}
        + \lambda_k (\thetab_1 - \thetab_2)
        \\
        & =  \sum_{\substack{k' < k, \\ h \le H}} \sum_{s' \in \Scal_{k',h}} \left( \nabla_{\thetab} p_{k',h}(s', \bar{\thetab}) \right)^\top (\thetab_1 - \thetab_2) \varphib_{k', h, s'}  + \lambda_k (\thetab_1 - \thetab_2)
        \\
        & =  \sum_{\substack{k' < k, \\ h \le H}} \left[\Hb_{k'} + \lambda_k \Ib_d\right](\thetab_1 - \thetab_2) \, ,
    \end{align*}
    where 
    \begin{equation*}
        \Hb_{k'} := \sum_{s' \in \Scal_{k',h} } p_{k',h} (s', \thetab) \varphib_{k',h, s'} \varphib_{k', h, s'}^\top - \sum_{s' \in \Scal_{k',h} } \sum_{s'' \in \Scal_{k',h} } p_{k',h} (s', \thetab)  p_{k',h} (s'', \thetab) \varphib_{k',h, s'} \varphib_{k', h, s''}^\top \, .
    \end{equation*}
    Also, to identify that $\Hb_{k'}$ is positive semi-definite, note that 
    \begin{equation*}
        (\xb - \yb)(\xb - \yb)^\top = \xb \xb^\top + \yb \yb^\top -\xb \yb^\top - \yb \xb^\top \succcurlyeq \zero_d
    \end{equation*}
    which implies $\xb \xb^\top + \yb \yb^\top \succcurlyeq \xb \yb^\top + \yb \xb^\top$ for any $\xb, \yb \in \RR^{d}$. 
    Therefore, it follows that
    \begin{align*}
        \Hb_{k'} 
        & = \sum_{s' \in \Scal_{k',h} } p_{k',h} (s', \thetab) \varphib_{k',h, s'} \varphib_{k', h, s'}^\top - \sum_{s' \in \Scal_{k',h} } \sum_{s'' \in \Scal_{k',h} } p_{k',h} (s', \thetab)  p_{k',h} (s'', \thetab) \varphib_{k',h, s'} \varphib_{k', h, s''}^\top 
        \\
        & = \sum_{s' \in \Scal_{k',h} } p_{k',h} (s', \thetab) \varphib_{k',h, s'} \varphib_{k', h, s'}^\top 
        \\
        & \phantom{{}={}} - \frac{1}{2} \sum_{s' \in \Scal_{k',h} } \sum_{s'' \in \Scal_{k',h} } p_{k',h} (s', \thetab)  p_{k',h} (s'', \thetab) (\varphib_{k',h, s'} \varphib_{k', h, s''}^\top + \varphib_{k',h, s''} \varphib_{k', h, s'}^\top) 
        \\
        & \succcurlyeq \sum_{s' \in \Scal_{k',h} } p_{k',h} (s', \thetab) \varphib_{k',h, s'} \varphib_{k', h, s'}^\top 
        \\
        & \phantom{{}={}}- \frac{1}{2} \sum_{s' \in \Scal_{k',h} } \sum_{s'' \in \Scal_{k',h} } p_{k',h} (s', \thetab)  p_{k',h} (s'', \thetab) (\varphib_{k',h, s'} \varphib_{k', h, s'}^\top + \varphib_{k',h, s''} \varphib_{k', h, s''}^\top) 
        \\
        & = \sum_{s' \in \Scal_{k',h} } p_{k',h} (s', \thetab) \varphib_{k',h, s'} \varphib_{k', h, s'}^\top - \sum_{s' \in \Scal_{k',h} } \! \sum_{s'' \in \Scal_{k',h} } p_{k',h} (s', \thetab) p_{k',h} (s'', \thetab) \varphib_{k',h, s'} \varphib_{k', h, s'}^\top 
        \\
        & = \sum_{s' \ne \dot{s}_{k',h}} p_{k',h} (s', \thetab) \varphib_{k',h, s'} \varphib_{k', h, s'}^\top 
            - \sum_{s' \ne \dot{s}_{k',h}} \sum_{s'' \ne \dot{s}_{k',h}  } p_{k',h} (s', \thetab) p_{k',h} (s'', \thetab) \varphib_{k',h, s'} \varphib_{k', h, s'}^\top
        \\
        & = \sum_{s' \ne \dot{s}_{k',h}} p_{k',h} (s', \thetab) \left( 1 - \sum_{s'' \ne \dot{s}_{k',h}} p_{k',h} (s'', \thetab) \right) \varphib_{k',h, s'} \varphib_{k', h, s'}^\top
        \\
        & \succcurlyeq  \sum_{s' \ne \dot{s}_{k',h}} \kappa \varphib_{k',h, s'} \varphib_{k', h, s'}^\top \, ,
    \end{align*}
    where the last inequality comes from the Assumption~\ref{asm_kappa}.
    Hence, for any $\thetab \in \RR^{d}$, since $\Gb_k(\thetab^*) = \zero_d$, we have
    \begin{align*}
        \| \Gb_k (\thetab) \|_{\Ab_k^{-1}}^2 
        & = \| \Gb_k (\thetab) - \Gb_k(\thetab^*) \|_{\Ab_k^{-1}}^2
        \\
        & = \left( \Gb_k (\thetab) - \Gb_k(\thetab^*) \right)^\top \Ab_k^{-1} \left( \Gb_k (\thetab) - \Gb_k(\thetab^*) \right)
        \\
        & = (\thetab - \thetab^*)^\top \left( \sum_{\substack{k' < k, \\ h \le H}} \Hb_{k'} + \lambda \Ib_d \right) \Ab_k^{-1}
        \left( \sum_{\substack{k' < k, \\ h \le H}} \Hb_{k'} + \lambda_k \Ib_d \right) (\thetab - \thetab^*)
        \\
        & \ge \kappa^2 (\thetab - \thetab^*)^\top \Ab_k (\thetab - \thetab^*)
        \\
        & = \kappa^2 \| \thetab - \thetab^* \|_{\Ab_k}^2
    \end{align*}
    where the inequality follows from the fact:
    \begin{equation*}
        \sum_{\substack{k' < k, \\ h \le H}} \Hb_{k'} + \lambda_k \Ib_d \succcurlyeq \kappa \Ab_k \, .
    \end{equation*}
    
    Now, from the definition of $\Gb_k (\hat{\thetab}_k)$, we have
    \begin{align*}
        \kappa \| \hat{\thetab}_k - \thetab^* \|_{\Ab_k}
        & \le \| \Gb_k (\hat{\thetab}_k) \|_{\Ab_k^{-1}}
        \\
        & \le \left\| \sum_{\substack{k' < k, \\ h \le H}} \sum_{s' \in \Scal_{k',h}} \epsilon_{k',h}^{s'} \varphib_{k',h, s'} \right\|_{\Ab_k^{-1}} + \lambda_k \|  \thetab^* \|_{\Ab_k^{-1}} \, .
    \end{align*}
    
    For the next step, we introduce the following lemma.
    \begin{lemma}[Lemma 1 in~\citet{li2024improved}] \label{lemma:self-normalized martingale with correlated noise}
        Let $\{ \Fcal_t \}_{t=0}^{\infty}$ be a filtration, and $\{ \zetab_t \}_{t=1}^{\infty}$ be an $\RR^N$-valued stochastic process such that $\zetab_t$ is $\Fcal_t$-measurable one-hot vector.
        Suppose that $\EE[\deltab_t \mid \Fcal_{t-1}] = \pb_t$ and define $\epsilonb_t = \pb_t - \deltab_t$.
        Let $\{ \Xb_t = [\xb_{t,1}^\top, \cdots, \xb_{t,N}^\top]^\top \}_{t=1}^\infty$ be a sequence of $\RR^{N \times d}$-valued stochastic process such that $\Xb_{t}$ is $\Fcal_{t-1}$-measurable and $\| \xb_{t,j} \|_2 \le 1$, for all $j \in [N]$.
        Let $\{ \lambda_t \}_{t=1}^\infty$ be a sequence of non-negative scalars.    
        Define
        \begin{equation*}
            \Yb_t = \sum_{i=1}^t \sum_{j=1}^N \xb_{i,j} \xb_{i,j}^\top + \lambda_t \Ib_d \, ,
            \quad
            \sbb_t = \sum_{i=1}^t \sum_{j=1}^N \epsilonb_{t,j} \xb_{i,j} \, .
        \end{equation*}
        Then, for any $\delta \in (0,1)$, with probability $1 - \delta$, we have for all $t \ge 1$,
        \begin{equation*}
            \| \sbb_t \|_{\Yb_t^{-1}} \le \frac{\sqrt{\lambda_t}}{4} + \frac{4}{\sqrt{\lambda_t}} \log \left( \frac{2^d \det (\Yb_t)^{1/2} \lambda_t^{-d/2}}{\delta} \right) \, .
        \end{equation*}
    \end{lemma}
    
    Note that for $k' \in [k-1], h \in [H]$, $\epsilon^{s'}_{k',h} = y^{s'}_{k',h} - p_{k',h}(s', \thetab^*)$ and $[y^{s'}_{k',h}]_{s' \in \Scal_{k',h}}$ is an one-hot vector.
    Then by, Lemma~\ref{lemma:self-normalized martingale with correlated noise}, with probability $1-\delta$, we have for all $k \ge 1$,
    \begin{align*}
        \left\| \sum_{\substack{k' < k, \\ h \le H}} \sum_{s' \in \Scal_{k',h}} \epsilon_{k',h}^{s'} \left( \frac{\varphib_{k',h, s'}}{L_{\varphib}} \right) \right\|^2_{\tilde{\Ab}_k^{-1}}
        \le \frac{ L_{\varphib} \sqrt{\lambda_k}}{4} + \frac{4}{L_{\varphib} \sqrt{\lambda_k}} \log \left( \frac{2^d \det(\tilde{\Ab}_k)^{1/2} (L_{\varphib}^2 \lambda_k)^{-d/2}}{\delta} \right) \, ,
    \end{align*}
    where $L_{\varphib}$ is the $\ell_2$-norm upper bound of the feature $\varphib_{k,h,s'}$ and $\tilde{\Ab}_k$ is defined as
    \begin{equation*}
        \tilde{\Ab}_k
        = \sum_{\substack{k'<k \\ h \le H}} \sum_{s' \in \Scal_{k',h}} \left(\frac{\varphib_{k',h, s'}}{L_{\varphib}}\right) \left(\frac{\varphib_{k',h, s'}}{L_{\varphib}}\right)^\top + L_{\varphib}^2 \lambda_k \Ib_d \, .
    \end{equation*}
    Since $\tilde{\Ab}_k = \frac{1}{L_{\varphib}^2} \Ab_k$, we have
    \begin{align*}
        \left\| \sum_{\substack{k' < k, \\ h \le H}} \sum_{s' \in \Scal_{k',h}} \epsilon_{k',h}^{s'} \left( \frac{\varphib_{k',h, s'}}{L_{\varphib}} \right) \right\|^2_{\tilde{\Ab}_k^{-1}}
        & = \left\| \sum_{\substack{k' < k, \\ h \le H}} \sum_{s' \in \Scal_{k',h}} \epsilon_{k',h}^{s'} \varphib_{k',h, s'} \right\|^2_{\Ab_k^{-1}}
        \\
        & \le \frac{ L_{\varphib} \sqrt{\lambda_k}}{4} + \frac{4}{L_{\varphib} \sqrt{\lambda_k}} \log \left( \frac{(2/L_{\varphib}^2)^d \det(\Ab_k)^{1/2} \lambda_k^{-d/2}}{\delta} \right)
        \\
        & \le \frac{L_{\varphib} \sqrt{ \lambda_k}}{4} + \frac{4}{L_{\varphib} \sqrt{\lambda_k}} \left( d \log \frac{2}{L_{\varphib}^2} + \frac{d}{2} \log \left( 1 + \frac{k H \Ucal L_{\varphib}^2}{d \lambda_k} \right) - \log \delta \right) \, ,
    \end{align*}
    where the second inequality comes from the determinant-trace inequality (Lemma~\ref{lemma:det-tr ineq}).
    
    On the other hand, for $ \| \thetab^* \|_{\Ab_k^{-1}}$, we have
    \begin{equation*}
         \| \thetab^* \|_{\Ab_k^{-1}}^2
         = {\thetab^*}^\top \Ab_k^{-1} \thetab^*
         \le \frac{1}{\lambda_k} {\thetab^*}^\top \thetab^*
         = \frac{\| \thetab^* \|_2^2}{\lambda_k} \, .
    \end{equation*}
    Hence, we have $\lambda_k \| \thetab^* \|_{\Ab_k^{-1}} \le \sqrt{\lambda_k} L_{\thetab}$.
    Then, by combining the results, with probability at least $1-\delta$, we have for all $k \ge 1$, 
    \begin{align*}
        \| \hat{\thetab}_k - \thetab^* \|_{\Ab_k}
        & \le \underbrace{ \kappa^{-1} \left( \bigg( \frac{L_{\varphib}}{4} + L_{\thetab}  \bigg) \sqrt{\lambda_k}
         + \frac{4}{L_{\varphib}\sqrt{\lambda_k}} \left( d \log \frac{2}{L_{\varphib}^2} + \frac{d}{2} \log \left( 1 + \frac{k H \Ucal L_{\varphib}^2}{d \lambda_k} \right) - \log \delta \right) \right)}_{=:\beta_k(\delta)} \, .
    \end{align*}
\end{proof}

\subsection{Proof of Lemma~\ref{lemma_optimism of estimated q function}}
\begin{proof}[Proof of Lemma~\ref{lemma_optimism of estimated q function}]
    We prove this by backwards induction on $h$. 
    For the base case $h = H$, by definition of~\eqref{eq:UC Q-function}
    \begin{equation*}
        \hat{Q}_{k, H} (s,a) = r(s,a) = Q^*_H (s,a)
    \end{equation*}
    since $\hat{V}_{k, H + 1}(s) = V^*_{H+1}(s) = 0$ for all $s \in \Scal$.
    Suppose that the statement holds for step $h+1$ for some $1 \le h \le H-1$.
    Then, for step $h$ and for all $(s,a) \in \Scal \times \Acal$,
    \begin{align*}
        \hat{Q}_{k, h} (s,a) 
        & = r(s,a) + \max_{\thetab \in \Ccal_k}  \frac{\sum_{s' \in \Scal_{s,a}} \exp(\varphib(s,a, s')^\top \thetab ) \hat{V}_{k,h+1}(s')}{\sum_{s' \in \Scal_{s,a}} \exp(\varphib(s,a,s')^\top \thetab)}
        \\
        & \ge r(s,a) +  \frac{\sum_{s' \in \Scal_{s,a}} \exp(\varphib(s,a, s')^\top \thetab^* ) \hat{V}_{k,h+1}(s')}{\sum_{s' \in \Scal_{s,a}} \exp(\varphib(s,a, s')^\top \thetab^* )}
        \\
        & = r(s,a) + P_h \hat{V}_{k, h+1}(s,a)
        \\
        & \ge r(s,a) + P_h V^*_{h+1}(s,a)
        \\
        & = Q^*_{h}(s,a)
    \end{align*}
    where the first inequality holds by the construction of $\hat{Q}_{k,h}$, the second inequality follows from the induction hypothesis.
\end{proof}

\subsection{Proof of Lemma~\ref{lemma:value f.t concentration}}
\begin{proof}[Proof of Lemma~\ref{lemma:value f.t concentration}]
    Let 
    \begin{equation*}
        \tilde{\thetab} := \argmax_{\thetab \in \Ccal_t}  \frac{\sum_{s' \in \Scal_{k,h}} \exp(\varphib_{k,h, s'}^\top \thetab ) \hat{V}_{k,h+1}(s')}{\sum_{s' \in \Scal_{k,h}} \exp(\varphib_{k,h, s'}^\top \thetab )} \, .
    \end{equation*}
    Then, we have
        \begin{align}
            & \hat{Q}_{k,h} (s_{k,h}, a_{k,h}) - \left[ r(s_{k,h}, a_{k,h}) + P_h \hat{V}_{k, h+1} (s_{k,h}, a_{k,h}) \right] \nonumber
            \\ 
            & =  \frac{\sum_{s' \in \Scal_{k,h}} \exp(\varphib_{k,h,s'}^\top \tilde{\thetab} ) \hat{V}_{k,h+1}(s')}{\sum_{s' \in \Scal_{k,h}} \exp(\varphib_{k,h, s'}^\top \tilde{\thetab} )} 
            -  \frac{\sum_{s' \in \Scal_{k,h}} \exp(\varphib_{k,h,s'}^\top \thetab^* ) \hat{V}_{k,h+1}(s')}{\sum_{s' \in \Scal_{k,h}} \exp(\varphib_{k,h, s'}^\top \thetab^* )} \label{eq:value f.t concen 1} \, .
        \end{align}
    Then by the mean value theorem, there exists $\bar{\thetab}=\nu \tilde{\thetab} + (1-\nu) \thetab^*$ for some $0 \le \nu \le 1$ satisfying that
        \begin{align*}
            (\ref{eq:value f.t concen 1})
            & = \frac{ \left( \sum_{s' \in \Scal_{k,h}}  \exp( \varphib_{k,h,s'}^\top \bar{\thetab}) \hat{V}_{k,h+1}(s') \varphib_{k,h,s'}^\top (\tilde{\thetab} - \thetab^*) \right) \left( \sum_{s' \in \Scal_{k,h}} \exp( \varphib_{k,h,s'}^\top \bar{\thetab}) \right) }{ \left( \sum_{s' \in \Scal_{k,h}} \exp( \varphib_{k,h,s'}^\top \bar{\thetab}) \right)^2}
            \\
            & \phantom{{}={}}
            - \frac{ \left( \sum_{s' \in \Scal_{k,h}} \exp( \varphib_{k,h,s'}^\top \bar{\thetab}) \hat{V}_{k,h+1}(s') \right) \left( \sum_{s' \in \Scal_{k,h}} \exp( \varphib_{k,h,s'}^\top \bar{\thetab}) \varphib_{k,h,s'}^\top (\tilde{\thetab} - \thetab^*) \right)}{ \left( \sum_{s' \in \Scal_{k,h}} \exp( \varphib_{k,h,s'}^\top \bar{\thetab}) \right)^2}
            \\
            & = \sum_{s' \in \Scal_{k,h}} p_{k,h}(s', \bar{\thetab}) \hat{V}_{k,h+1}(s') \varphib_{k,h,s'}^\top (\tilde{\thetab} - \thetab^*)
            \\
            & \phantom{{}={}}
            - \left(  \frac{\sum_{s' \in \Scal_{k,h}} \exp( \varphib_{k,h,s'}^\top \bar{\thetab}) \hat{V}_{k,h+1}(s')}{\sum_{s' \in \Scal_{k,h}} \exp( \varphib_{k,h,s'}^\top \bar{\thetab})} \right)
            \sum_{s' \in \Scal_{k,h}} p_{k,h}(s', \bar{\thetab}) \varphib_{k,h,s'}^\top (\tilde{\thetab} - \thetab^*)
            \\
            & = \sum_{s' \in \Scal_{k,h}} \left( \hat{V}_{k,h+1}(s') -  \frac{\sum_{s' \in \Scal_{k,h}} \exp( \varphib_{k,h,s'}^\top \bar{\thetab}) \hat{V}_{k,h+1}(s')}{\sum_{s' \in \Scal_{k,h}} \exp( \varphib_{k,h,s'}^\top \bar{\thetab})} \right)
            p_{k,h}(s', \bar{\thetab}) \varphib_{k,h,s'}^\top (\tilde{\thetab} - \thetab^*)
            \\
            & \le H \sum_{s' \in \Scal_{k,h}} p_{k,h}(s', \bar{\thetab}) \varphib_{k,h,s'}^\top (\tilde{\thetab} - \thetab^*)
            \\
            & \le H \max_{s' \in \Scal_{k,h}} | \varphib_{k,h,s'}^\top (\tilde{\thetab} - \thetab^*) |
            \\
            & \le H \max_{s' \in \Scal_{k,h}} \| \varphib_{k,h,s'}\|_{\Ab_k^{-1}} \| \tilde{\thetab} - \thetab^* \|_{\Ab_k}
            \\
            & \le H \max_{s' \in \Scal_{k,h}} \| \varphib_{k,h,s'}\|_{\Ab_k^{-1}} ( \| \tilde{\thetab} - \hat{\thetab}_k \|_{\Ab_k} + \|\hat{\thetab}_k - \thetab^* \|_{\Ab_k})
            \\
            & \le 2 H \beta_k (\delta) \max_{s' \in \Scal_{k,h}} \| \varphib_{k,h,s'}\|_{\Ab_k^{-1}} \, .
        \end{align*}
    where the second inequality holds due to the definition of $p_{k,h}(s', \bar{\thetab})$.
\end{proof}
\subsection{Proof of Theorem~\ref{thm_main theorem}}
Now we are ready to provide the proof of Theorem~\ref{thm_main theorem}.
\begin{proof}[Proof of Theorem~\ref{thm_main theorem}]
    By the condition of $\lambda_k$, we can guarantee that $\| \varphib_{k,h,s'} \|_{\Ab_k^{-1}}$ is smaller than 1 for all $k \in [K], h \in [H]$ and $s' \in \Scal_{k,h}$, which means,
    \begin{equation*}
        \| \varphib_{k,h,s'} \|^2_{\Ab_k^{-1}}
        = \varphib_{k,h,s'}^\top \Ab_k^{-1} \varphib_{k,h,s'}
        \le \frac{1}{\lambda_k} \| \varphib_{k,h,s'} \|_2^2
        \le \frac{1}{\lambda_k} L_{\varphib}^2 
        \le 1 \, .
    \end{equation*}
    Let $\delta' = \delta / 2$.
    Then by Lemma~\ref{lemma:draft of regret bound}, with probability at least $1 - \delta/2$, we have 
    \begin{align*}
        \Regret(K) 
        & \le 2 H \beta_K (\delta'/K) \sqrt{K H \sum_{k=1}^K \sum_{h=1}^H \max_{s' \in \Scal_{k,h}} \| \varphib_{k,h,s'} \|_{\Ab_k^{-1}}^2 }
        + \sum_{k=1}^K \sum_{h=1}^H \dot{\zeta}_{k,h}
        \\
        & \le 2 H \beta_K (\delta'/K) \sqrt{2 T H d \log \left( 1 + K H \Ucal L_{\varphib}^2 \right) }
        + \sum_{k=1}^K \sum_{h=1}^H \dot{\zeta}_{k,h} \, ,
    \end{align*}
    where the second inequality follows from the Lemma~\ref{lemma:sum of weighted norm}.
    On the other hand, note that $\dot{\zeta}_{k,h}$ is a martingale difference satisfying $| \dot{\zeta}_{k,h} | \le 2 H$ and $\EE[\dot{\zeta}_{k,h} \mid \Fcal_{k,h}] = 0$ where $\Fcal_{k,h}:= \{ s_{k',h}, a_{k',h}, y_{k',h}:k' < k, h \le H \} \cup \{ s_{k,h'}, a_{k,h'}, y_{k, h'-1} : h' \le h \}$. 
    By applying Azuma-Hoeffiding inequality, with probability at least $1 - \delta/2$, we have
    \begin{equation*}
        \sum_{k=1}^K \sum_{h=1}^H \dot{\zeta}_{k,h} \le H \sqrt{2T \log(2/\delta)} \, .
    \end{equation*}
    By taking union bound, with probability at least $1 - \delta$, we get the desired result as follows:
    \begin{equation*}
        \Regret(K) = \tilde{\Ocal} \left( \kappa^{-1} d \sqrt{H^3 T} \right) \, .
    \end{equation*}
\end{proof}

\section{Supporting Lemmas Used in Appendix~\ref{appx_proof of thm1}}\label{appx:supporting lemmas}
In this section, we provide some supporting lemmas used in Appendix~\ref{appx_proof of thm1}.

\begin{lemma}[Good event] \label{lemma:good event}
    For $\delta' \in (0,1)$, define the following event:
    \begin{equation*}
        \Ecal_{k}(\delta') = \left\{ \| \hat{\thetab}_{k'} - \thetab^* \|_{\Ab_{k'}} \le \beta_{k'} (\delta'), \forall k' \le k \right\} \, .
    \end{equation*}
    Then, for any $K \in \NN$, the good event $\Ecal_{K}(\delta')$ holds with probability at least $1 - \delta$ where $\delta' = \delta / K$.
\end{lemma}
\begin{proof}[Proof of Lemma~\ref{lemma:good event}]
    By Lemma~\ref{lemma_theta hat concentration}, the event $\Ecal_K(\delta')$ happens with probability at least $1 - \delta$, where $\delta' = \delta / K$.
\end{proof}

\begin{lemma} \label{lemma:draft of regret bound}
    For $\delta \in (0,1)$, with probability $1 - \delta$, 
    we have
    \begin{equation*}
        \Regret(K) 
        \le 2 H \beta_K (\delta/K) \sqrt{K H \sum_{k=1}^K \sum_{h=1}^H \max_{s' \in \Scal_{k,h}} \| \varphib_{k,h,s'} \|_{\Ab_k^{-1}}^2 }
        + \sum_{k=1}^K \sum_{h=1}^H \dot{\zeta}_{k,h} \, ,
    \end{equation*}
    where $\dot{\zeta}_{k,h} =  \EE_{s' | s_{k,h}, a_{k,h}} \Big[ \left( \hat{V}_{k,h+1} - V^{\pi_k}_{h+1} \right)(s') \Big] - \left( \hat{V}_{k,h+1} - V^{\pi_k}_{h+1} \right)(s_{k,h+1})$.
\end{lemma}

\begin{proof}[Proof of Lemma~\ref{lemma:draft of regret bound}]
    By Lemma~\ref{lemma:good event}, the good event $\Ecal_K(\delta')$ holds with probability at least $1 - \delta$ where $\delta' = \delta/K$. 
    Conditioned on the event $\Ecal_K(\delta')$, by Lemma~\ref{lemma_optimism of estimated q function}, we have
    \begin{align*}
        ( V^*_1 - V^{\pi_k}_1) (s_{k,1})
        & = ( Q^*_1 (s_{k,1}, \pi^* (s_{k,1})) - V^{\pi_k}_1 (s_{k,1}))
        \\
        & \le ( \hat{Q}_{k,1} (s_{k,1}, \pi^* (s_{k,1})) - V^{\pi_k}_1 (s_{k,1}))
        \\
        & \le ( \hat{Q}_{k,1} (s_{k,1}, a_{k,1}) - V^{\pi_k}_1 (s_{k,1})) \, .
    \end{align*}
    On the other hand, by Lemma~\ref{lemma:value f.t concentration}, we have
    \begin{equation*}
        \hat{Q}_{k,h} (s_{k,h}, a_{k,h}) 
        \le 2 H \beta_k (\delta') \max_{s' \in \Scal_{k,h}} \| \varphib_{k,h,s'} \|_{\Ab_k^{-1}}
        + \left[ r(s_{k,h}, a_{k,h}) + P_h \hat{V}_{k, h+1} (s_{k,h}, a_{k,h}) \right] \, .
    \end{equation*}
    Then we have
    \begin{align*}
        \hat{Q}_{k,1} (s_{k,1}, a_{k,1}) - V^{\pi_k}_1 (s_{k,1})
        & = \hat{Q}_{k,1} (s_{k,1}, a_{k,1}) - Q^{\pi_k}_1 (s_{k,1}, a_{k,1}) 
        \\
        & \le 2 H \beta_k (\delta') \max_{s' \in \Scal_{k,1}} \| \varphib_{k,1,s'} \|_{\Ab_k^{-1}}
        + P_1 ( \hat{V}_{k,2} - V^{\pi_k}_{2} )(s_{k,1}, a_{k,1}) 
        \\
        & = 2 H \beta_k (\delta') \max_{s' \in \Scal_{k,1}} \| \varphib_{k,1,s'} \|_{\Ab_k^{-1}}
        + \underbrace{ \EE_{s' | s_{k,1}, a_{k,1}} \Big[ \left( \hat{V}_{k,2} - V^{\pi_k}_{2} \right)(s') \Big] }_{ \dot{\zeta}_{k,1} + \left( \hat{V}_{k,2} - V^{\pi_k}_{2} \right)(s_{k,2}) } 
        \\
        & = 2 H \beta_k (\delta') \max_{s' \in \Scal_{k,1}} \| \varphib_{k,1,s'} \|_{\Ab_k^{-1}}
        + \left( \hat{V}_{k,2} - V^{\pi_k}_{2} \right)(s_{k,2}) + \dot{\zeta}_{k,1}
    \end{align*}
    where the inequality comes from the Lemma~\ref{lemma:value f.t concentration}.
    Applying this logic recursively, we have
    \begin{align*}
        \hat{Q}_{k,1} (s_{k,1}, a_{k,1}) - V^{\pi_k}_1 (s_{k,1})
        & \le \sum_{h=1}^{H} 2 H \beta_k (\delta') \max_{s' \in \Scal_{k,h}} \| \varphib_{k,h,s'} \|_{\Ab_k^{-1}}
        + \sum_{h=1}^{H} \dot{\zeta}_{k,h} 
    \end{align*}
    where $\dot{\zeta}_{k,h} =  \EE_{s' | s_{k,h}, a_{k,h}} \Big[ \left( \hat{V}_{k,h+1} - V^{\pi_k}_{h+1} \right)(s') \Big] - \left( \hat{V}_{k,h+1} - V^{\pi_k}_{h+1} \right)(s_{k,h+1})$.
    
    By summing up over all episodes, we get the desired result.
    \begin{align*}
        \Regret(K)
        & = \sum_{k=1}^K ( V^*_1 - V^{\pi_k}_1 ) (s_{k,1})
        \\
        & \le 2 H \beta_K (\delta') \sum_{k=1}^K \sum_{h=1}^H \max_{s' \in \Scal_{k,h}} \| \varphib_{k,h,s'} \|_{\Ab_k^{-1}}
        + \sum_{k=1}^K \sum_{h=1}^H \dot{\zeta}_{k,h}
        \\
        & \le 2 H \beta_K (\delta') \sqrt{KH \sum_{k=1}^K \sum_{h=1}^H \max_{s' \in \Scal_{k,h}} \| \varphib_{k,h,s'} \|_{\Ab_k^{-1}}^2}
        + \sum_{k=1}^K \sum_{h=1}^H \dot{\zeta}_{k,h} \, .
    \end{align*}
\end{proof}
\begin{lemma} \label{lemma:sum of weighted norm}
    Define $\Ab_k = \lambda \Ib_d + \sum_{\substack{k'<k \\ h \le H}} \sum_{s' \in \Scal_{k',h}}\! \varphib_{k',h, s'} \varphib_{k',h, s'}^\top$. 
    If $\lambda \ge L_{\varphib}^2 $, then
    \begin{equation*}
        \sum_{\substack{k'<k \\ h \le H}} \max_{s' \in \Scal_{k',h}}  \| \varphib_{k',h, s'} \|_{\Ab_k^{-1}}^2 
        \le
        2 H d \log \left( 1 + \frac{k H \Ucal L_{\varphib}^2}{d \lambda} \right) \, .
    \end{equation*}
\end{lemma}

\begin{proof}
    Recall that
    \begin{equation*}
        \Ab_{k+1} = \Ab_k + \sum_{h \le H} \sum_{s' \in \Scal_{k,h}} \varphib_{k,h,s'} \varphib_{k,h,s'}^\top \, . 
    \end{equation*}
    Since we have $\Ab_k \succ \zero$, then we have
    \begin{equation*}
        \det \Ab_{k+1} = \det (\Ab_k)  \det \left( I_d + \Ab_k^{-1/2} \sum_{h=1}^H \sum_{s' \in \Scal_{k,h}} \varphib_{k,h,s'} \varphib_{k,h,s'}^\top \Ab_k^{-1/2} \right) \, .
    \end{equation*}
    Let $\lambda_i$ be an eigenvalue of $I_d + \Ab_k^{-1/2} \sum_{h=1}^H \sum_{s' \in \Scal_{k,h}} \varphib_{k,h,s'} \varphib_{k,h,s'}^\top \Ab_k^{-1/2}$, $1 \le i \le d$. Then, for all $1 \le i \le d$, we have $\lambda_i \ge 1$ and 
    \begin{equation*}
        \sum_{i=1}^{d} (\lambda_i - 1) = \tr \left(I_d + \Ab_k^{-1/2} \sum_{h=1}^H \sum_{s' \in \Scal_{k,h}} \varphib_{k,h,s'} \varphib_{k,h,s'}^\top \Ab_k^{-1/2} \right) - d = \sum_{h=1}^H \sum_{s' \in \Scal_{k,h}} \| \varphib_{k,h,s'} \|_{\Ab_k^{-1}}^2
    \end{equation*}
    Therefore, we have
    \begin{align*}
        \det\left(I_d + \Ab_k^{-1/2} \sum_{h=1}^H \sum_{s' \in \Scal_{k,h}} \varphib_{k,h,s'} \varphib_{k,h,s'}^\top \Ab_k^{-1/2}\right)
        & = \prod_{i=1}^{d} \lambda_i
        \\
        & = \prod_{i=1}^{d} \left[ 1 + (\lambda_i - 1 ) \right]
        \\
        & \ge 1 + \sum_{i=1}^{d} (\lambda_i - 1 )
        \\
        & = 1 + \sum_{h=1}^H \sum_{s' \in \Scal_{k,h}} \| \varphib_{k,h,s'} \|_{\Ab_k^{-1}}^2
        \\
        & \ge 1 + \sum_{h=1}^H \max_{s' \in \Scal_{k,h}} \| \varphib_{k,h,s'} \|_{\Ab_k^{-1}}^2
    \end{align*}
    where the first inequality comes from the fact that $\left[1 + (a-1)\right] \left[1 + (b-1)\right] = 1 + (a-1) + (b-1) + (a-1)(b-1) \ge 1 + (a-1) + (b-1)$ for some $a,b \ge 1$.
    On the other hand, 
    \begin{align*}
        1 + \sum_{h=1}^H \max_{s' \in \Scal_{k,h}} \| \varphib_{k,h,s'} \|_{\Ab_k^{-1}}^2 
        & = \frac{\sum_{h=1}^H ( 1 + H \max_{s' \in \Scal_{k,h}} \| \varphib_{k,h,s'} \|_{\Ab_k^{-1}}^2) }{H}
        \\
        & \ge \prod_{h=1}^H ( 1 + H \max_{s' \in \Scal_{k,h}} \| \varphib_{k,h,s'} \|_{\Ab_k^{-1}}^2)^{1/H}
        \\
        & \ge \prod_{h=1}^H ( 1 + \max_{s' \in \Scal_{k,h}} \| \varphib_{k,h,s'} \|_{\Ab_k^{-1}}^2)^{1/H} \, ,
    \end{align*}
    where the first inequality comes from the AM–GM inequality.
    Then, it follows that
    \begin{align*}
        \det \Ab_{k+1} 
        & \ge \det (\Ab_k) \prod_{h=1}^H ( 1 + \max_{s' \in \Scal_{k,h}} \| \varphib_{k,h,s'} \|_{\Ab_k^{-1}}^2)^{1/H}
        \\
        & \ge \det (\lambda \Ib_d) \prod_{k'=1}^k \prod_{h=1}^H ( 1 + \max_{s' \in \Scal_{k',h}} \| \varphib_{k',h,s'} \|_{A_{k'}^{-1}}^2)^{1/H} \, .
    \end{align*}
    Since $\lambda \ge L_{\varphib}^2 $, we have for all $k \in [K], h \in [H]$ and $s' \in \Scal_{k,h}$
   \begin{equation*}
        \| \varphib_{k,h,s'} \|^2_{\Ab_k^{-1}}
        = \varphib_{k,h,s'}^\top \Ab_k^{-1} \varphib_{k,h,s'}
        \le \frac{1}{\lambda} \| \varphib_{k,h,s'} \|_2^2
        \le \frac{1}{\lambda} L_{\varphib}^2
        \le 1 \, .
    \end{equation*}
    Then, using the fact that $z \le 2\log(1+z)$ for any $z \in [0,1]$, we have
    \begin{align*}
         \sum_{\substack{k'<k \\ h \le H}} \max_{s' \in \Scal_{k',h}}  \| \varphib_{k',h, s'} \|_{A_{k'}^{-1}}^2 
        & \le 2 \sum_{\substack{k'<k \\ h \le H}} \log \left( 1 + \max_{s' \in \Scal_{k',h}}  \| \varphib_{k',h, s'} \|_{A_{k'}^{-1}}^2  \right)
        \\
        & = 2 \log \prod_{k'=1}^k \prod_{h=1}^H \left( 1 + \max_{s' \in \Scal_{k',h}}  \| \varphib_{k',h, s'} \|_{A_{k'}^{-1}}^2  \right)
        \\
        & \le 2 H \log \left( \frac{\det \Ab_{k+1}}{\det \lambda \Ib_d} \right)
        \\
        & \le 2 H \log \left[ \left( \frac{\tr(\lambda \Ib_d) + k H \Ucal L_{\varphib}^2}{d} \right)^{d} \frac{1}{\det \lambda \Ib_d} \right]
         \\
         &
         = 2 H d \log \left( 1 + \frac{k H \Ucal L_{\varphib}^2}{d \lambda} \right) \, ,
    \end{align*}
    where the last inequality follows from the Lemma~\ref{lemma:det-tr ineq}.
\end{proof}

\section{Limitations}~\label{appx:limitations}
We assume a particular parametric model, the MNL model for the transition model of MDPs. Hence, realizability is assumed. Note that such realizability assumption is also imposed in almost all previous literature on provable reinforcement learning with function approximation~\citet{yang2020reinforcement, jin2020provably, zanette2020frequentist, modi2020sample, du2020is, cai2020provably, ayoub2020model, wang2020reinforcement_eluder, weisz2021exponential, he2021logarithmic, zhou2021nearly, zhou2021provably, ishfaq2021randomized}. However, we hope that this condition can be relaxed in the future work.

\section{Additional Lemmas}

\begin{lemma}[Determinant-Trace Inequality~\citet{abbasi2011improved}]\label{lemma:det-tr ineq}
    Suppose $\xb_1, \ldots, \xb_t \in \RR^d$ and for any $1 \le s \le t$, $\| \xb_s \|_2 \le L$.
    Let $\bar{\Vb}_t = \lambda \Ib_d + \sum_{s=1}^t \xb_s \xb_s^\top$ for some $\lambda > 0$.
    Then,
    \begin{equation*}
        \det \bar{\Vb}_t \le (\lambda + t L^2 /d)^d \, .
    \end{equation*}
\end{lemma}

\begin{lemma}[Lemma 8 in~\citet{oh2019thompson}]~\label{lemma:lemma 8 in Oh et al. (2019)}
    Suppose $\| \xb_{t,i} \|_2 \le 1$ for all $t \le T, 1 \le i \le K$. Define $\Vb_t = \lambda \Ib_d + \sum_{\tau=1}^t \sum_{i = 1}^K \xb_{\tau, i} \xb_{\tau, i}^\top$. 
    If $\lambda \ge 1$, then
    \begin{equation}
        \sum_{\tau=1}^t \max_{i \in [K]}  \| \xb_{t,i} \|_{\Vb_t^{-1}}^2 
        \le
        2 \log \left( \frac{\det \Vb_t }{\lambda^{d}} \right) \, .
    \end{equation}
\end{lemma}

\section{Experiment Details}~\label{appx:experiment details}

 \begin{figure*}
        \centering
        \begin{subfigure}[b]{0.45\textwidth}
            \centering
            \includegraphics[scale=0.40]{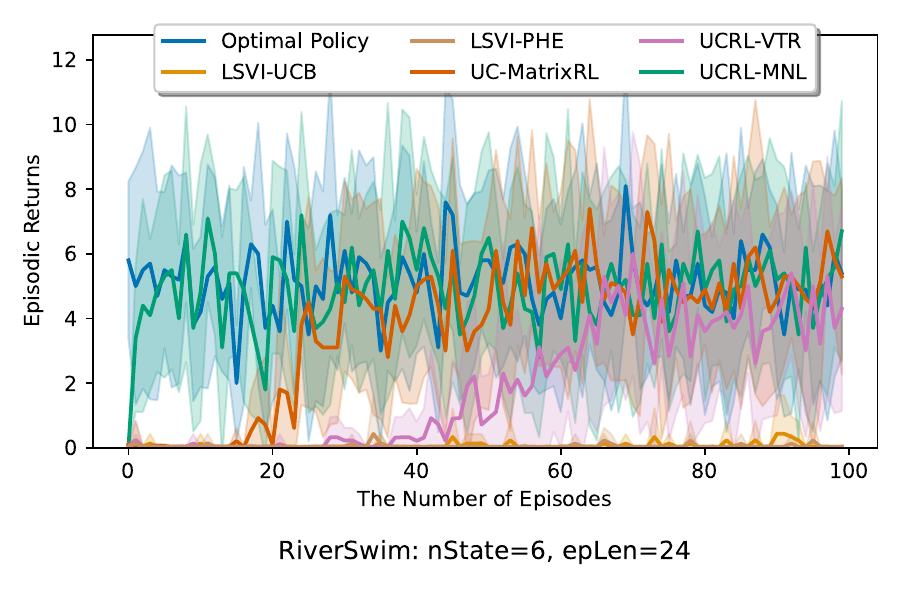}
            \label{fig:RiverSwim S=6, H=24}
        \end{subfigure}
        \begin{subfigure}[b]{0.45\textwidth}  
            \centering 
            \includegraphics[scale=0.40]{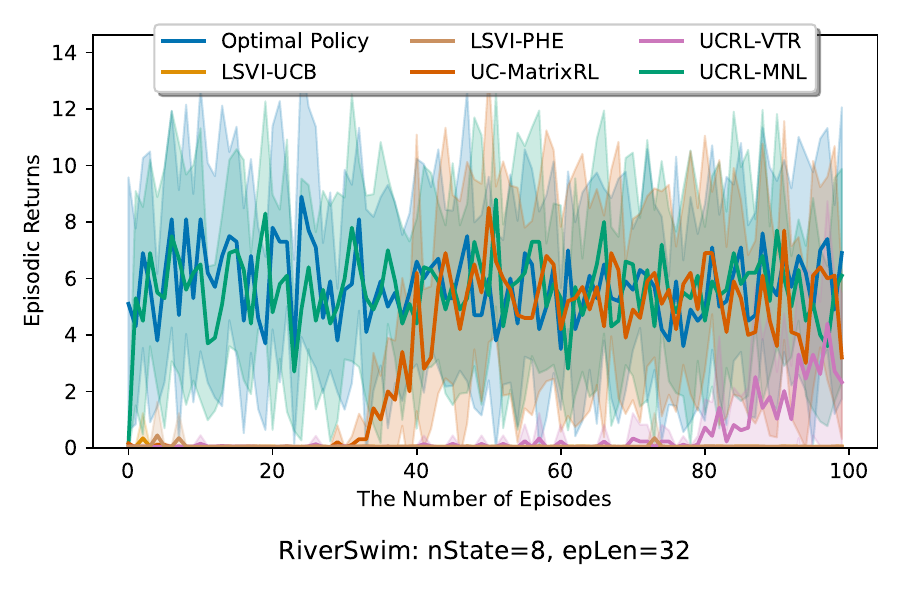}
            \label{fig:RiverSwim S=8, H=32}
        \end{subfigure}
        \begin{subfigure}[b]{0.45\textwidth}   
            \centering 
            \includegraphics[scale=0.40]{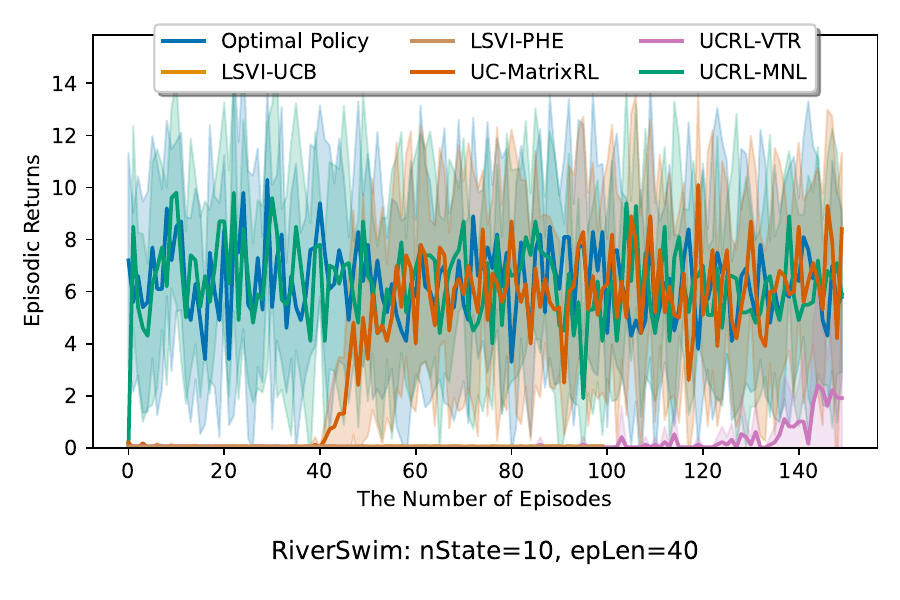}
            \label{fig:RiverSwim S=10, H=40}
        \end{subfigure}
        \begin{subfigure}[b]{0.45\textwidth}   
            \centering 
            \includegraphics[scale=0.40]{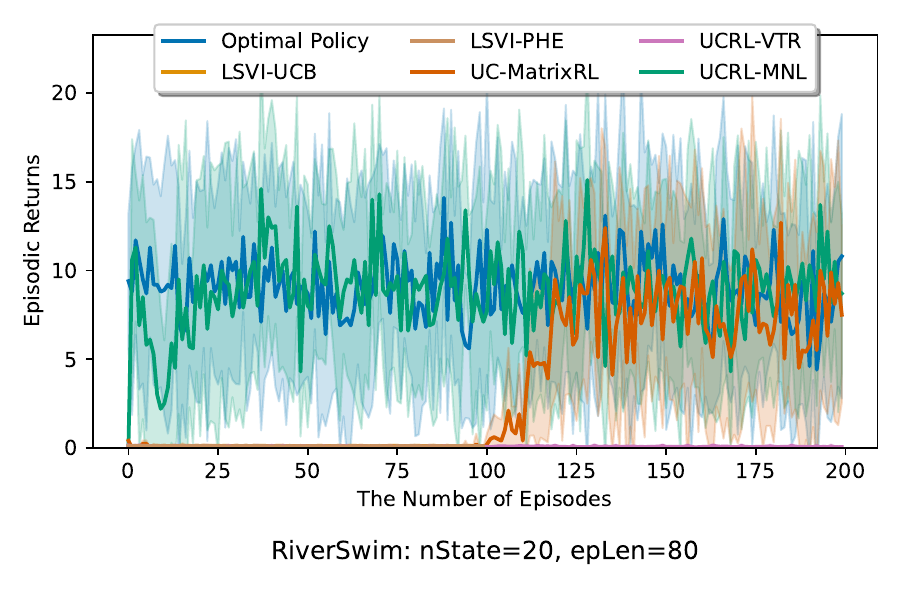}
            \label{fig:RiverSwim S=20, H=80}
        \end{subfigure}
        \caption[ The average and standard deviation of critical parameters ]
        {Episodic returns over 10 independent runs under the different RiverSwim environments (without smoothing effect)} 
        \label{fig:RiverSwim_ns}
\end{figure*}

In this section, we will explain the result for the RiverSwim in more detail.
In Figure~\ref{fig:RiverSwim}, the smoothing effect was applied for the sake of clear visualization of overall tendency. 
However, such smoothing may overshadow the performance of our algorithm in the early episodes --- since it can make the results seem as if our proposed algorithm reached the optimal policy from the beginning. 
The episodic return without smoothing effect is presented in Figure~\ref{fig:RiverSwim_ns}. 
As can be seen in Figure~\ref{fig:RiverSwim_ns}, our proposed algorithm converges to the optimal policy within a \textbf{few} episodes.
Compared to the benchmark algorithms, the reason why our proposed algorithm is able to converge quickly is that at every time step, other algorithms use only one feedback about the state where the actual transition occurred, whereas the proposed algorithm uses information of reachable states including the state in which no actual transition is occurred.
This means that the proposed algorithm can get more information from one sample.
Also, if the size of the set of reachable states is properly bounded (Note that $\Ucal = 3$ in this experiment) even when the size of the state space grows, we can see that our proposed algorithm has superior performance consistently, independent of the size of the state space.

\end{document}